\documentclass{article}
\usepackage{arxiv}

\usepackage{amssymb} 
\usepackage{graphicx} 

\usepackage{amsmath}
\usepackage{booktabs}
\usepackage{multirow}
\usepackage{threeparttable}

\usepackage{xcolor}
\usepackage{listings}

\lstdefinestyle{awamjson}{
    basicstyle=\ttfamily\footnotesize,
    backgroundcolor=\color{black!3},
    frame=single,
    rulecolor=\color{black!20},
    framerule=0.4pt,
    framesep=6pt,
    xleftmargin=7pt,
    xrightmargin=7pt,
    breaklines=true,
    breakatwhitespace=false,
    columns=fullflexible,
    keepspaces=true,
    showstringspaces=false,
    tabsize=2,
    aboveskip=0pt,
    belowskip=0pt
}

\usepackage{wrapfig}
\usepackage{algorithm}
\usepackage{algpseudocode}
\usepackage{float}
\usepackage{xcolor}

\floatstyle{ruled}
\restylefloat{algorithm}

\usepackage{booktabs}
\usepackage{tabularx}
\usepackage{caption}

\usepackage{pifont}

\usepackage{soul}
\usepackage{xcolor}
\usepackage[utf8]{inputenc} 
\usepackage[T1]{fontenc}    
\usepackage{hyperref}       
\usepackage{url}            
\usepackage{booktabs}       
\usepackage{amsfonts}       
\usepackage{nicefrac}       
\usepackage{microtype}      
\usepackage{lipsum}		
\usepackage{graphicx}
\usepackage{doi}
\usepackage{multirow}
\usepackage{multicol}
\usepackage{xcolor}
\usepackage{subcaption}
\usepackage[normalem]{ulem}
\usepackage{amsmath,amsfonts,bm}
\usepackage{amssymb,amsthm}
\usepackage{enumitem}
\usepackage{subcaption}
\usepackage{adjustbox}
\usepackage{afterpage}
\usepackage[
    backend=biber,
    citestyle=authoryear,
    bibstyle=authortitle,
    mincitenames=1,
    maxcitenames=1,
    maxbibnames=3
]{biblatex}
\newcommand{\ourname}{{AWAM-0:}}

\newcommand{\citep}[1]{\parencite{#1}}
\renewcommand{\thefootnote}{\fnsymbol{footnote}}
\setlist[itemize,1]{leftmargin=\dimexpr 18pt}
\setlist[enumerate,1]{leftmargin=\dimexpr 18pt}

\title{%
\bfseries
\raisebox{-0.1\height}{%
\includegraphics[width=0.036\textwidth]{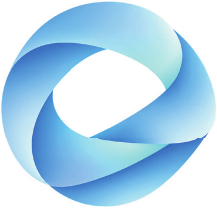}} %
\ourname~Agentic World Action Model for\\
Embodied Intelligence
}

\author{%
\normalfont\normalsize\bfseries
Ruiqi Song*, Baiyong Ding*, Shimin Shu, Chenming Zhang,
Ruilin Wang, Dujun Nie, Yuanhui Huang,\\[5pt]\normalfont\normalsize\bfseries
Lei Fan, Yuchen Li, Wenzhao Zheng, Yuhang He,
Chen Lv, Hangbin Wu, Long Chen\textdagger
}

\newcommand{\InstitutionLogos}{%
  \noindent\makebox[\textwidth][c]{%
    \includegraphics[width=0.82\textwidth]{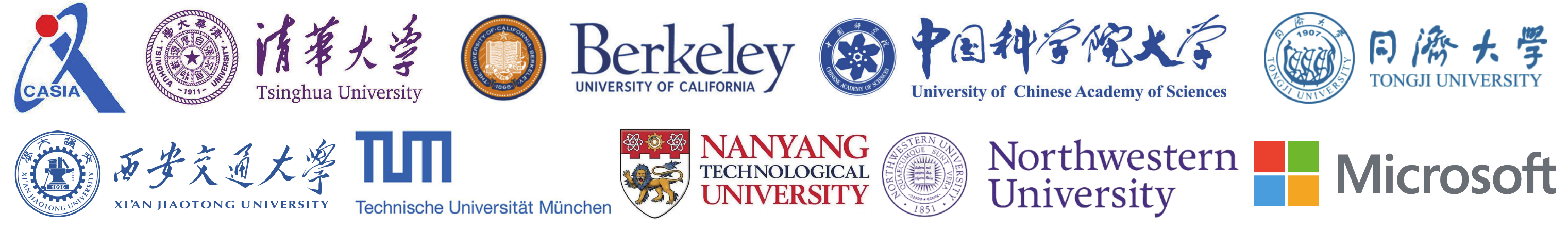
    }%
  }\par
}

\date{}

\renewcommand{\headeright}
{\raisebox{-0.18\height}{\includegraphics[width=0.024\textwidth]{figs/mobi.pdf}}~OpenSpace-Lab~Technical Report}
\renewcommand{\undertitle}{}
\renewcommand{\shorttitle}

\begin{document}
\maketitle

\vspace{-30pt}
\InstitutionLogos
\vspace{20pt}
\begin{abstract}
Existing Vision-Language-Action (VLA) models and World Action Models (WAMs) typically rely on end-to-end decision-making architectures without explicit agentic reasoning. Although WAMs provide foresight through future prediction, they lack agentic closed-loop reasoning and remain vulnerable to error accumulation in long-horizon, compositional, and out-of-distribution tasks. To address this limitation, we introduce AWAM, an Agentic World Action Model whose dual-brain architecture comprises an Agentic Reasoning Brain and a World–Action Brain, integrating high-level reasoning, world modeling, and action generation. The Agentic Reasoning Brain reasons over task goals, observations, and retrieved experience to generate and refine structured intents. Conditioned on these intents, the World–Action Brain jointly predicts future world states and executable actions, distilling the predicted physical consequences of these actions into world evidence that is fed back to the Agentic Reasoning Brain. This interaction establishes a unified closed loop in which reasoning and world modeling constrain and refine each other to guide action generation. During training, AWAM generates multiple candidate intents conditioned on the same physical state and evaluates them through simulation rollouts. At inference time, AWAM adopts a single-intent execution strategy and re-deliberates when necessary. Experiments across multiple benchmarks demonstrate AWAM’s superior manipulation performance, with strong capabilities in long-horizon execution, compositional generalization, out-of-distribution robustness, and failure recovery.
\end{abstract}


\begingroup
\renewcommand{\thefootnote}{}
\footnotetext{%
    {{{*} Equal Contribution}}
\qquad {{\textdagger Corresponding author}}
}
\endgroup

\begingroup
\renewcommand{\thefootnote}{}
\footnotetext{%
    {Project page:}
    \url{https://ruiqi-song.github.io/AWAM/}
}
\endgroup


\begin{figure*}[h]
    \centering
        \includegraphics[width=0.9\textwidth]{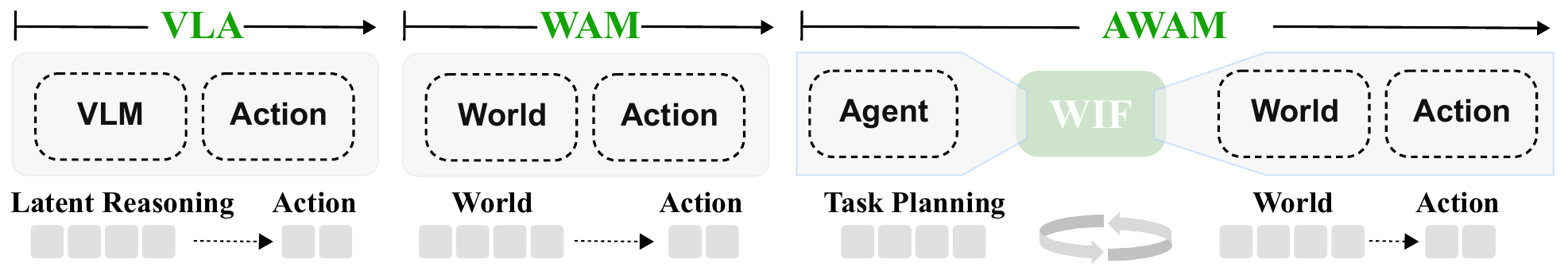}
   \caption{Comparison of VLA, WAM, and AWAM architectures. AWAM couples agentic reasoning with world action brain via a dual-brain closed loop. WIF denotes World Imagination and Feedback.}
    \vspace{-12pt}
    \label{fig:overview}
\end{figure*}
\section{Introduction}

Recent Vision--Language--Action models (VLAs), such as RT-2 and OpenVLA, use vision--language pretraining and diverse robot data to support general-purpose embodied control~\citep{brohan2023rt2,kim2024openvla}. However, most VLAs map observations and instructions directly to actions without explicitly modeling action consequences or maintaining persistent task states. This limits their ability to handle long-horizon dependencies, skill composition, and environmental perturbations. In such settings, accumulated errors can lead to task-stage confusion and task failure.

World Action Models (WAMs) extend this paradigm by jointly modeling future world states and robot actions, thereby learning physical priors that support action generation~\citep{ye2026dreamzero,wang2026wam}. These learned representations can improve action prediction even without explicit video generation at inference time~\citep{yuan2026fastwam}. However, world modeling provides foresight but not agency. Predicting action consequences alone does not inherently enable WAMs to identify futures worth exploring, compare strategies, or revise decisions based on execution feedback. World modeling alone therefore does not provide the agentic closed-loop reasoning needed to maintain global objectives, resolve cross-stage dependencies, and recover from failures.

Recent approaches augment frozen VLAs or WAMs with large language model agents for task decomposition, skill orchestration, and experience retrieval~\citep{zhang2026harnessvla,gu2026harnesswam}. However, these systems largely rely on external orchestration, coupling agents and policies through textual subgoals or primitive invocations while providing only coarse execution feedback to the agents. Such interfaces limit the bidirectional exchange of high-level intents and fine-grained physical feedback. Moreover, interaction experience is primarily stored in external memory rather than internalized by the underlying model to improve its physical capabilities.

This progression highlights a missing closed loop among reasoning, world modeling, and action: reasoning should identify futures worth imagining, while world modeling should assess their physical feasibility. To this end, we introduce AWAM, a dual-brain architecture comprising an Agentic Reasoning Brain and a World--Action Brain. The former generates structured physical intents from task goals, observations, and retrieved experience. Conditioned on these intents, the latter jointly predicts future world states and executable actions and feeds back physical evidence to refine high-level decisions. This bidirectional interaction unifies reasoning, world modeling, and action generation within a coherent closed loop.
The model's evolution is illustrated in Fig.~\ref{fig:overview}.

We further develop a lightweight simulator-grounded reinforcement learning procedure. 
After supervised initialization, the frozen Agent proposes multiple
primitive intents, and the old policy collects independent rollouts
from the same initial state. Execution returns optimize the Action Head through FPO++, while eligible outcomes supervise the Physical Head. The video branch remains frozen, and persistent memory stays
fixed. Candidate execution feedback guides subsequent intent generation. The proposed inference procedure reviews a single intent before manipulation and uses review and execution feedback for replanning.
Experiments across multiple benchmarks demonstrate consistent improvements
over existing approaches, highlighting the effectiveness of AWAM.
Our main contributions are summarized as follows:
\begin{itemize}
\item We introduce AWAM, a dual-brain architecture that couples a frozen Agentic Reasoning Brain with a World--Action Brain through structured intents and physical feedback, integrating
high-level reasoning, world prediction, and action generation.

\item We develop an online reinforcement learning procedure based on multi-intent rollouts, using simulator returns to optimize
the Action Head and auxiliary physical supervision to update
the Physical Head, while keeping the reasoning backbone
and video branch frozen.

\item We construct a Task--Skill--Primitive data pipeline that converts robot demonstrations into hierarchical task knowledge,
primitive-conditioned action supervision, and physical labels, supporting memory construction and supervised world--action learning.

\item We evaluate AWAM on RoboDojo, BiCoord, and real-world robotic manipulation tasks, assessing task completion, long-horizon execution, generalization, and failure recovery.

\end{itemize}

\section{Related Work}


\subsection{Vision--Language--Action Models}
Vision--Language--Action models have progressed from multi-task policies toward generalist control through scalable robot learning and the transfer of Internet-scale vision--language knowledge~\citep{brohan2022rt1,brohan2023rt2}. Cross-embodiment data aggregation, large-scale pretraining, and efficient adaptation have further broadened their applicability across robots and tasks~\citep{oneill2023openx,octo2024,kim2024openvla}. More recently, flow-based generative modeling of continuous action chunks has enabled fine-grained control and modeling of multimodal action distributions~\citep{black2024pi0}. Despite these advances in data scaling, knowledge transfer, and action modeling, VLAs primarily map observations and instructions directly to actions, with limited capacity to explicitly model future physical consequences, maintain persistent task states, or perform closed-loop replanning.

\subsection{World Action Models}
World Action Models extend direct action prediction by jointly modeling future world states and robot actions. Explicit video prediction captures action-conditioned environmental dynamics, while large-scale video-generation priors help models learn transferable physical dynamics~\citep{ye2026dreamzero}. Recent developments increasingly emphasize control-oriented world representations through action-relevant future modeling and latent-space transition prediction~\citep{gigaworld2026,beingh07_2026}. Integrating predictive knowledge during training can further reduce reliance on costly explicit video generation at inference time~\citep{yuan2026fastwam}. These advances improve physical foresight and data utilization. Nevertheless, existing WAMs remain centered on finite-horizon world--action prediction, with limited support for agentic reasoning, strategy deliberation, and feedback-driven decision revision to achieve global objectives.

\subsection{Agent-Augmented World Action Models}
Integrating agents with robotic policies initially relied on language models to compose predefined skills, executable programs, or spatial value fields, linking semantic planning with low-level control~\citep{ahn2022saycan,liang2023codepolicies,huang2023voxposer}. As local manipulation capabilities improved, this integration expanded to task decomposition, experience memory, execution monitoring, and failure recovery, with task-state maintenance and world evidence supporting long-horizon decisions~\citep{zhang2026harnessvla,gu2026harnesswam}. However, these approaches largely rely on system-level orchestration through task graphs, textual subgoals, and execution feedback, leaving reasoning and world--action modeling loosely coupled. AWAM instead conditions world--action modeling on structured physical intents and returns predicted physical consequences to the Agentic Reasoning Brain, enabling bidirectional dual-brain collaboration.

\section{Method}
\subsection{Problem Formulation}
\label{sec:problem_formulation}
\begin{figure*}[t]
    \centering
    \includegraphics[width=0.9\textwidth]{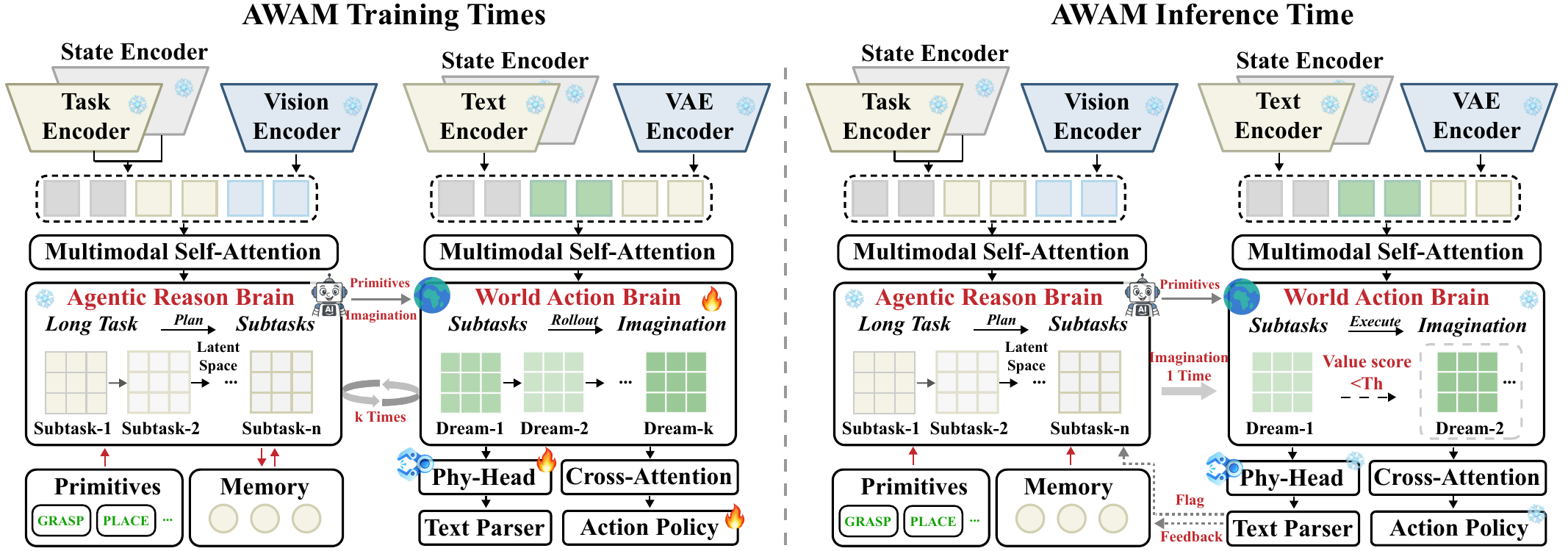}
\caption{Overview of AWAM.
Left: Multi-intent rollout training updates the Action and Physical Heads using execution returns and physical supervision,
with the video branch frozen and persistent memory fixed.
Right: Inference retrieves experience to generate a single intent
by default, adaptively refining it with world evidence and
execution feedback.}
    \label{fig:awam_overview}
    \vspace{-12pt}
\end{figure*}

\textbf{Interaction Data.}~
We define the decision context as $x_t=(g,o_t,h_t,m_t)$,
where $g$ denotes the task instruction, $o_t$ the current
observation and robot state, $h_t$ the interaction history,
and $m_t$ the experience retrieved from memory $\mathcal{M}_t$.
During online training, the frozen Agent proposes $K$ structured
intents $\{z_t^{j,k}\}_{k=1}^{K}$ in each of $J$ deliberation
rounds. Each WAM intent yields one execution trajectory
$\tau_t^{j,k}$ from the same simulator snapshot and a local
success label $r_t^{j,k}\in\{0,1\}$.
These records form
$\mathcal{D}=\{(x_t,z_t^{j,k},\tau_t^{j,k},r_t^{j,k})\}$.
For each recorded action chunk, $N$ flow-time/noise probes
are cached for FPO++ updates.



\textbf{Learning Objective.}~
We aim to improve full-task success by adapting the parameters
$\theta$ of the World--Action Brain $\pi_W$ and updating experience
memory while keeping the Agentic Reasoning Brain $\pi_A$ frozen.
Let $\mathbf{Z}=(z_0,\ldots,z_{L-1})$ denote the intents executed
over $L$ decision steps and $\mathcal{T}$ the full trajectory
composed of segments $\tau_t$.
We express the probability
to maximize as follows:
\begin{equation}
\label{eq:task_success_probability}
p_{\theta,\mathcal{M}_0}(R=1\mid g)
=
\int p(R=1\mid g,\mathcal{T})
\prod_{t=0}^{L-1}
\pi_{A}(z_t\mid x_t)
P_{\theta}(\tau_t\mid x_t,z_t)
\,d\mathbf{Z}\,d\mathcal{T}
\end{equation}
Here, $\mathcal{M}_0$ is the initial memory, and $P_\theta$ is
the trajectory distribution induced by $\pi_W$ and the environment.
The factorization assumes that $x_t$ summarizes relevant history
and memory, with rollout evidence and memory-update variables
jointly marginalized.
Adapting $\pi_W$ aims to improve intent-conditioned execution,
while updated memory guides subsequent intents through $m_t$
with $\pi_A$ frozen.
Local labels $r_t^{j,k}$ provide outcome supervision,
whereas $R$ indicates full-task success.
Fig.~\ref{fig:awam_overview} shows the overall model architecture.


\textbf{Discussion.}~
Existing approaches connect frozen language-model agents with
pretrained VLAs or WAMs through subgoal interfaces and execution
feedback~\citep{zhang2026harnessvla,gu2026harnesswam}.
The central challenge is to connect high-level reasoning with fine-grained physical evidence and translate interaction experience into improved execution performance.
Successful rollouts reveal effective behaviors, while failed rollouts expose limitations in proposed intents and their execution.
This motivates our central question:
\emph{Can predicted physical consequences and accumulated
experience iteratively refine high-level reasoning and improve
long-horizon task success while preserving efficient action
generation from world representations?}

\subsection{Recursive Reason--World Interaction} \label{sec:reason_world_deliberation} 
\begin{wrapfigure}{r}{0.33\columnwidth}
    \centering
    \includegraphics[width=\linewidth]{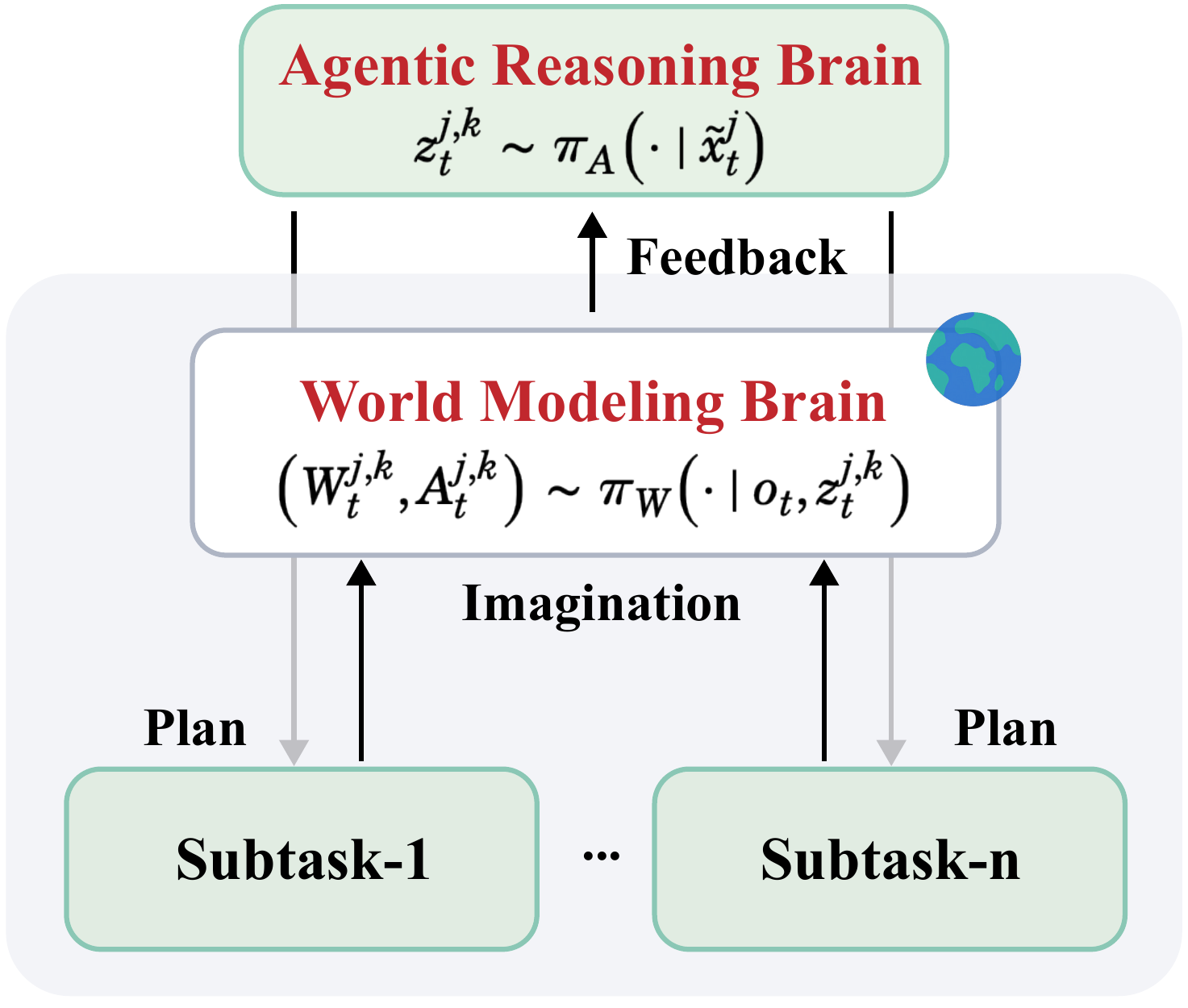}
    \caption{The recurrent reason--world deliberation across subtasks.}
    \label{fig:reason_world_deliberation}
    \vspace{-24pt}
\end{wrapfigure}

AWAM couples high-level reasoning with predictive world modeling
to generate executable actions, as shown in Fig.~\ref{fig:reason_world_deliberation}.
During online training, the decision context $x_t$ remains
fixed over $J$ deliberation rounds. In round $j$, the Agentic
Reasoning Brain proposes $K$ candidate intents
$\{z_t^{j,k}\}_{k=1}^{K}$.
Each WAM intent conditions future prediction and action
generation, yielding one execution rollout from the same
simulator snapshot. Execution feedback from round $j$
guides intent generation in round $j+1$.
During inference, the Agent proposes one intent per
planning attempt. Its predicted future is reviewed before
action generation, with feedback guiding replanning
when needed.

\subsubsection{Agentic Reasoning Brain}
\label{sec:agentic_brain}
The Agentic Reasoning Brain uses a frozen Qwen3.5~\citep{qwen2026qwen35} to track long-horizon task progress,
decompose remaining objectives into ordered subtasks, and select
a primitive for the active subtask. Its reasoning is conditioned on the task context and physical
evidence from the World Brain.

At step $t$ and round $j$, we define
the reasoning context as $\tilde{x}_t^j=(x_t,\mathbf{e}_t^{j-1})$,
combining the decision context with world evidence from the
previous round.
Let $S=(s_1,\ldots,s_n)$ denote the remaining subtask sequence,
with $s_1$ active, and let $c=(p,\eta)$ denote an invocation
of primitive $p\in\mathcal{P}$ with arguments $\eta$.
The frozen Agentic Reasoning Brain $\pi_A$ factorizes its
reasoning distribution as follows:
\begin{equation}
\label{eq:agentic_reasoning}
\pi_{A}(S,c\mid\tilde{x}_t^j)
=
\left[
\prod_{\ell=1}^{n}
\pi_{A}(s_\ell\mid\tilde{x}_t^j,s_{<\ell})
\right]
\pi_{A}(c\mid\tilde{x}_t^j,S)
\end{equation}
The first factor generates dependent subtasks autoregressively, while the second selects the active primitive and its arguments given the plan.
Eq.~\ref{eq:agentic_reasoning} groups generated tokens by subtask,
including termination in the final group.
World evidence $\mathbf{e}_t^{j-1}$ guides revisions to the remaining
plan and current primitive invocation.

The primitive library $\mathcal{P}$ includes controller-based
primitives for reset, recovery, and observation, and WAM-based
primitives for manipulation. The Agentic Reasoning Brain decomposes
the task into primitive-level intents $z_t^{j,k}$, with each complete
primitive composition $c$ submitted through the unified
\texttt{execute\_primitive} interface. For manipulation intents,
the World--Action Brain predicts physical evolution and generates
the corresponding actions.

\subsubsection{World Action Brain}
\label{sec:world_brain}

The World-Action Brain predicts how the current environment may evolve conditioned on an intent proposed by the Agentic Reasoning Brain. Given the current observation $o_t$ and primitive-level intent $z_t^{j,k}$, it predicts a sequence of latent world states over a horizon of $H$ steps. Rather than merely reconstructing future images, these representations capture decision-relevant physical dynamics, including task outcomes, interaction states, physical risks, and environmental constraints.

For candidate $k$ at step $t$ and round $j$, it takes
$o_t$ and $z_t^{j,k}$ as input and predicts a latent
world-state sequence $\mathbf{w}_{t+1:t+H}^{j,k}$,
from which it extracts physical evidence
$\mathbf{e}_t^{j,k}$.
The joint distribution is as follows:
\begin{equation}
\label{eq:physical_world_modeling}
\begin{aligned}
p_{\mathrm W}(W,\mathbf{e}\mid o,z)
&=
p_{\phi}(W\mid o,z)\,
p_{\psi}(\mathbf{e}\mid W)
=
\left[
\prod_{h=1}^{H}
p_{\phi}(w_h\mid o,z,w_{<h})
\right]
p_{\psi}(\mathbf{e}\mid W)
\end{aligned}
\end{equation}
\begin{wrapfigure}{r}{0.33\columnwidth}
    \centering
    \includegraphics[width=\linewidth]{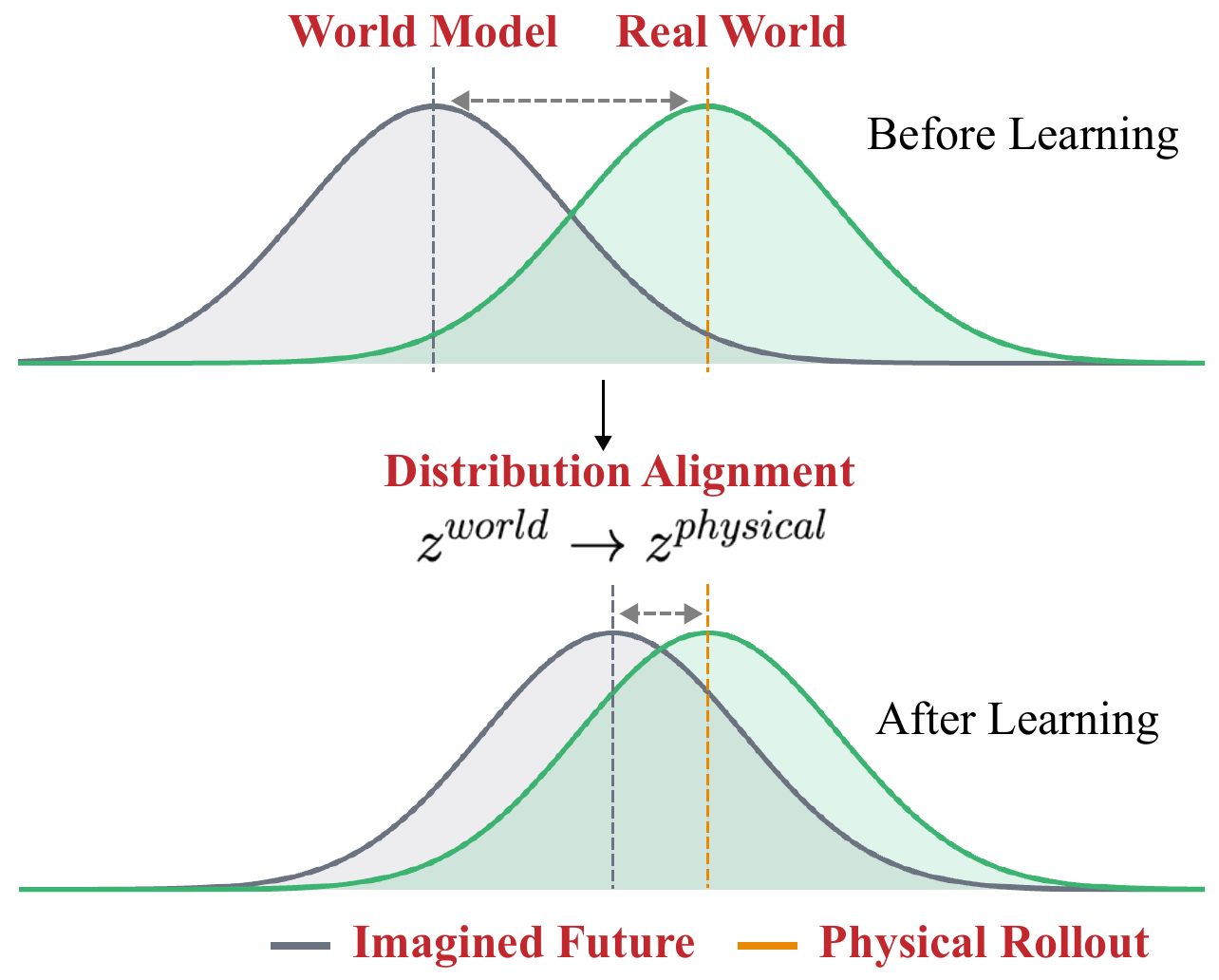}

    \caption{World--physical alignment before and after physical grounding.}
    \label{fig:distribution}
    \vspace{-36pt}
\end{wrapfigure}
Here, $p_{\phi}(W\mid o,z)$ models intent-conditioned physical evolution, while $p_{\psi}(\mathbf{e}\mid W)$ models physical evidence
including task outcomes, interaction states, and physical risks.
The dependency $z\rightarrow W\rightarrow\mathbf{e}$ ensures that intents are evaluated through predicted physical consequences rather than semantic reasoning alone.
Aggregated world states guide action denoising without directly passing $z$ to the Action Policy, while world evidence $\mathbf{e}$ refines the Agent's plan and primitive invocation. As illustrated in Fig.~\ref{fig:distribution}, multiple rounds of iteration progressively reduce the distribution gap between imagined futures and physical rollouts.

\subsubsection{Latent-Space Reason--World Recurrence}
\label{sec:reason_world_recurrence}
AWAM couples semantic intent reasoning with physical prediction
through recurrent evidence-guided refinement. At round $j$, let $\tilde{x}_t^j$ denote the reasoning context, including the task context, retrieved experience, and world
evidence from the preceding round.
Let $\mathcal{C}_t^j=\{z_t^{j,k}\}_{k=1}^{K}\subset\mathcal{Z}$
denote the candidate intent set.
The induced candidate prior $\pi_A(z\mid\tilde{x}_t^j)$ favors
intents consistent with the task and interaction history, but does not guarantee physical feasibility.

For each candidate $z$, the World-Action Brain predicts future world states and extracts physical evidence in terms of task outcomes, interaction states, and physical risks. A physical head estimates $p_W(y=1\mid o_t,z)$, where $y=1$
denotes successful intent execution.
Combining this estimate with the prior defines an idealized reason--world posterior:
\begin{equation}
\label{eq:reason_world_posterior}
\pi_{\mathrm{RW}}(z\mid\tilde{x}_t^j)
:=
p(z\mid\tilde{x}_t^j,y=1)
=
\frac{
\pi_A(z\mid\tilde{x}_t^j)\,
p_{\mathrm W}(y=1\mid o_t,z)
}{
\displaystyle
\sum_{z'\in\mathcal{C}_t^j}
\pi_A(z'\mid\tilde{x}_t^j)\,
p_{\mathrm W}(y=1\mid o_t,z')
}
\end{equation}
Here, $z\in\mathcal{C}_t^j$ denotes a candidate primitive intent, and $\pi_A$ represents the normalized prior weights from the Agentic Reasoning Brain. Eq.~4 provides a probabilistic reference
for balancing semantic preference and predicted execution success. During inference, AWAM applies fixed review rules to world evidence and returns review feedback to the Agent
for intent refinement. The reference posterior is not
explicitly computed.

\noindent\textbf{Proposition 1 (Variational Optimality of the
Reference Reason--World Posterior).}
\textit{
Let $\pi_A^j$ denote the candidate prior on $\mathcal{C}_t^j$,
and assume $\ell_W^j(z):=p_W(y=1\mid o_t,z)>0$ on its support.
Then the reference posterior in Eq.~\ref{eq:reason_world_posterior}
is the unique optimizer of the following objective:}
\begin{equation}
\label{eq:reason_world_variational}
\pi_{\mathrm{RW}}^j
=
\underset{
q\in\Delta(\mathcal{C}_t^j),\,q\ll\pi_A^j
}{
\operatorname{arg\,max}
}
\left[
\mathbb{E}_{z\sim q}\bigl[\log\ell_W^j(z)\bigr]
-
D_{\mathrm{KL}}(q\Vert\pi_A^j)
\right]
\end{equation}
Here, $q\ll\pi_A^j$ restricts $q$ to the prior's support. The first term favors intents with higher predicted success probabilities, while the second penalizes deviations from the semantic prior. This result characterizes the idealized reference posterior
without establishing the optimality of the fixed review rules or intent updates. A proof is provided in the Appendix.

The interaction becomes recurrent as feedback is incorporated
into the next reasoning context $\tilde{x}_t^{j+1}$.
The subsequent intent distribution is
$\pi_A^{j+1}(z):=\pi_A(z\mid\tilde{x}_t^{j+1})$.
The frozen Agent uses feedback to refine primitive invocations or retain successful intents. After deliberation, the world representation guides the Action Policy, while world evidence and retrieved experience refine reasoning and actions.

\subsection{World-Predictive Action Diffusion}
\label{sec:imagination_to_action}
AWAM transforms imagined physical futures into executable actions without
redundant world prediction. Since the imagined future remains fixed during
action generation, its representation is encoded once and reused throughout
denoising. The Physical Head evaluates imagined outcomes through
task outcomes, interaction states, and physical risks, providing world
evidence for selective deliberation and intent refinement.

\subsubsection{Diffusion Policy over Predictive World Latents}
\label{sec:efficient_action_diffusion}
Let $W_t$ denote the latent world-state sequence associated
with the selected intent.
We aggregate its multi-level representations into a future
context $\mathbf{f}_t=\mathcal{F}_{\omega}(W_t)$,
summarizing predicted physical evolution relevant to action generation.
Conditioned on the robot state $q_t$ and future context
$\mathbf{f}_t$, the Action Policy generates an action chunk
$A_t=(a_t,\ldots,a_{t+C-1})$ via reverse diffusion:
\begin{equation}
\label{eq:future_conditioned_action}
\begin{aligned}
p_{\theta_{\mathrm{act}}}(A_t^0\mid q_t,\mathbf{f}_t)
&=
\int p(A_t^D)
\prod_{d=1}^{D}
p_{\theta_{\mathrm{act}}}
(A_t^{d-1}\mid A_t^d,q_t,\mathbf{f}_t)
\,\mathrm{d}A_t^{1:D},\quad
A_t^D \sim\mathcal{N}(0,I),
A_t^0=A_t
\end{aligned}
\end{equation}
Here, $D$ is the number of denoising steps, $C$ is the action-chunk
length, and $\theta_{\mathrm{act}}$ denotes the action-denoising parameters.
The reverse process progressively transforms the initial noise
$A_t^D$ into executable actions $A_t^0$, reusing the same
future context $\mathbf{f}_t$ at every step.

\noindent\textbf{Reusable future context.}
During action denoising, the imagined future $W_t$ remains fixed
while action representations evolve.
AWAM computes $\mathbf{f}_t$ once and reuses it at selected
interaction layers, with the remaining layers performing
action-only refinement.
This reduces repeated world computation and dense world--action
interaction while retaining future information for action generation.

\noindent\textbf{Action generation from predicted futures.}
The Action Policy does not directly receive the latent intent $z_t$.
Instead, $z_t$ conditions the future world states predicted by
the World-Action Brain, and the resulting future context $\mathbf{f}_t$
is provided to the diffusion policy alongside the robot state $q_t$.


\subsubsection{Physical Evidence-Guided Action Inference}

Future-conditioned diffusion generates actions from imagined futures, but
high policy likelihood alone does not guarantee successful execution.
AWAM therefore evaluates imagined world-action hypotheses through the
Physical Head, which predicts structured physical evidence. AWAM predicts a single action--future pair per action chunk by default.
AWAM considers a single intent per planning attempt.
The proposed inference procedure reviews imagined outcomes before manipulation. Rejected or uncertain proposals return feedback to the Agent without execution. Approved proposals produce actions from the cached future context, while actual
execution feedback informs subsequent invocations.


\noindent\textbf{Shared Physical Evaluation.}
The structured world evidence $\mathbf{e}_t$ summarizes
predicted task outcomes, interaction states, and physical
risks. Fixed review rules use this evidence to assess the
proposed invocation, and the resulting feedback guides
the Agentic Reasoning Brain in plan refinement.
The same physical evaluation supports intent refinement
and execution decisions.

\noindent\textbf{Adaptive inference.}
AWAM defaults to $K=1$, avoiding candidate search and enabling efficient
future-conditioned action generation. When physical evidence indicates low
execution reliability or a mismatch between imagined and observed outcomes,
AWAM triggers an additional reason--world deliberation round. The Agentic
Reasoning Brain refines its plan and primitive invocation using world
evidence, after which the World--Action Brain evaluates the revised intent
before the Action Policy generates an updated action chunk. This maintains
efficient routine execution while enabling adaptive replanning when needed.

\subsection{Joint Learning of AWAM}
\label{sec:joint_learning}

AWAM is trained in two stages: supervised fine-tuning followed by online reinforcement learning. Throughout both stages, the Agentic Reasoning Brain remains frozen. Persistent memory stays fixed during rollout collection, while execution feedback updates the reasoning context.

\noindent\textbf{Stage I: Supervised Fine-Tuning.}
During supervised fine-tuning, the World--Action Brain and Physical Head
are jointly optimized using demonstration data. Given a target latent world-state
sequence $W$ and an action chunk $A$, we independently sample flow times
$\tau_{\mathrm W},\tau_{\mathrm A}\sim\mathcal{U}(0,1)$ and Gaussian noises
$\epsilon_{\mathrm W},\epsilon_{\mathrm A}$.
The interpolated states are defined as
$W_{\tau_{\mathrm W}}=(1-\tau_{\mathrm W})W+
\tau_{\mathrm W}\epsilon_{\mathrm W}$ and
$A_{\tau_{\mathrm A}}=(1-\tau_{\mathrm A})A+
\tau_{\mathrm A}\epsilon_{\mathrm A}$.
The corresponding target velocities are
$u_{\mathrm W}=\epsilon_{\mathrm W}-W$ and
$u_{\mathrm A}=\epsilon_{\mathrm A}-A$.
Let $\widehat{u}_{\mathrm W}$ and $\widehat{u}_{\mathrm A}$ denote the predicted velocities. The objectives are defined as follows:
\begin{equation}
\label{eq:wa_sft}
\begin{aligned}
\mathcal{L}_{\mathrm{WA}}
={}&
\lambda_{\mathrm W}
\mathbb{E}\left[
\omega_{\mathrm W}(\tau_{\mathrm W})
\left\|
\widehat{u}_{\mathrm W}-u_{\mathrm W}
\right\|_2^2
\right]
+
\lambda_{\mathrm A}
\mathbb{E}\left[
\omega_{\mathrm A}(\tau_{\mathrm A})
\left\|
\widehat{u}_{\mathrm A}-u_{\mathrm A}
\right\|_2^2
\right]
\end{aligned}
\end{equation}
The Physical Head predicts structured physical evidence from imagined world
states. Given physical labels
$Y_P=\{y_c\}_{c\in\mathcal{C}_{P}}$, its objective is as follows:
\begin{equation}
\label{eq:phy_sft}
\mathcal{L}_{\mathrm{PHY}}
=
-\sum_{c\in\mathcal{C}_{P}}
\lambda_c
y_c\log\hat y_c
\end{equation}
where $\mathcal{C}_{P}$ denotes the physical prediction categories of task
outcomes, interaction states, and physical risks. The supervised stage optimizes the World--Action Brain and Physical Head
with $\mathcal{L}_{\mathrm{SFT}}
=
\mathcal{L}_{\mathrm{WA}}
+
\lambda_{\mathrm{PHY}}\mathcal{L}_{\mathrm{PHY}} $.

\noindent\textbf{Stage II: Online Reinforcement Learning.}
After supervised initialization, AWAM performs online adaptation through
FPO++~\citep{yi2026flow}. During this stage, the future prediction component of the World--Action Brain is frozen, while the Action Head and Physical Head are updated from interaction feedback.
The old policy collects one execution rollout per candidate
WAM intent. An independent critic $V_\nu$ provides GAE
advantages, and $N$ shared flow-time/noise probes are cached
for each action chunk. Following the manipulation fine-tuning setting of FPO++, we use
per-sample surrogate ratios with PPO clipping:
\begin{equation}
\label{eq:fpo}
\mathcal{L}_{\mathrm{FPO++}}
=
-\mathbb{E}
\left[
\frac{1}{|\mathcal{B}|N}
\sum_{b\in\mathcal{B}}\sum_{i=1}^{N}
\min\!\left(
\rho_{\theta}^{b,i}\hat{A}^{b},
\operatorname{clip}\!\left(
\rho_{\theta}^{b,i},1-\epsilon,1+\epsilon
\right)\hat{A}^{b}
\right)
\right]
\end{equation}
Here, $\mathcal{B}$ is a mini-batch of cached action chunks,
$b\in\mathcal{B}$ indexes chunks, and $i=1,\ldots,N$ indexes
fixed flow-time/noise probes.
$\hat{A}^{b}$ is the chunk-start GAE advantage shared across
its probes, and $\epsilon$ is the PPO clipping parameter.
The surrogate ratio is
$\rho_{\theta}^{b,i}
=\exp(\ell_{\theta_{\mathrm{old}}}^{b,i}
-\ell_{\theta}^{b,i})$,
where $\ell_{\theta}^{b,i}$ sums the Stage-I action
flow-matching error over valid rows of chunk $b$, using
the same actions, conditions, and probes for both policies.
ActionDiT and the Physical Head use the following online
objective; the critic is trained separately against
GAE value targets:
\begin{equation}
\label{eq:online_objective}
\mathcal{L}_{\mathrm{online}}
=
\mathcal{L}_{\mathrm{FPO++}}
+
\lambda_{\mathrm{PHY}}'
\mathcal{L}_{\mathrm{PHY}}
\end{equation}
The two-stage optimization enables AWAM to first acquire general
world-action-physical representations from demonstrations and then
improve execution reliability through closed-loop interaction.

\begin{algorithm}[t]
\caption{Multi-Intent Rollout Training with FPO++}
\label{alg:multi_intent_training}
\small
\begin{algorithmic}[1]
\State \textbf{Input:} Frozen $\pi_A$;
$\pi_W$ with frozen video branch; critic $V_\nu$;
fixed memory $\mathcal{M}$;
$K,J,H,N,\gamma,\lambda_{\mathrm{GAE}},\epsilon,U$.
\For{each training iteration}
    \State Sample $(g,o_t,h_t)$; save snapshot $\sigma_t$;
    retrieve $m_t$ from $\mathcal{M}$
    \State $\pi_W^{\mathrm{old}}\leftarrow\pi_W$;
    $\nu_{\mathrm{old}}\leftarrow\nu$;
    $\mathcal{B}_t\leftarrow\varnothing$;
    $\mathbf{e}_t^0\leftarrow\varnothing$
    \For{$j=1,\ldots,J$}
        \hfill{\color{gray}\textit{// Multi-Intent Collection}}
        \State $\tilde{x}_t^j\leftarrow
        (g,o_t,h_t,m_t,\mathbf{e}_t^{j-1})$
        \State Generate $\{z_t^{j,k}\}_{k=1}^{K}$
        using $\pi_A(\cdot\mid\tilde{x}_t^j)$
        \For{each WAM intent $z_t^{j,k}$}
            \State Restore $\sigma_t$; collect $\tau_t^{j,k}$
            under $\pi_W^{\mathrm{old}}$ and $z_t^{j,k}$
            for up to $H$ steps
            \State Cache rewards, termination flags, old values,
            and $N$ probes with old losses per chunk
        \EndFor
        \State Collect $\mathcal{B}_t^j$;
        $\mathcal{B}_t\leftarrow\mathcal{B}_t\cup\mathcal{B}_t^j$
        \State $\mathbf{e}_t^j\leftarrow
        \operatorname{PublicFeedback}(\mathcal{B}_t^j)$
    \EndFor
    \State Compute GAE advantages $\widehat{A}$
    and value targets $\widehat{R}$ using $V_{\nu_{\mathrm{old}}}$
    \State Build eligible physical targets;
    fix data, probes, advantages, and targets
    \For{$u=1,\ldots,U$}
        \hfill{\color{gray}\textit{// Joint Optimization}}
        \For{each mini-batch $\mathcal{B}$}
            \State Compute block-summed $\ell_{\theta}^{b,i}$;
            $\rho_{\theta}^{b,i}\leftarrow
            \exp(\ell_{\theta_{\mathrm{old}}}^{b,i}
            -\ell_{\theta}^{b,i})$
            \State Compute $\mathcal{L}_{\mathrm{FPO++}}$
            using per-probe clipping and fixed
            chunk-start advantages $\widehat{A}^{b}$
            \State Compute $\mathcal{L}_{\mathrm{PHY}}$
            and $\mathcal{L}_{V}
            =\frac{1}{2}\operatorname{MSE}_{\mathrm{valid}}
            (V_\nu,\widehat{R})$
            \State $\mathcal{L}_{\mathrm{online}}\leftarrow
            \mathcal{L}_{\mathrm{FPO++}}
            +\lambda'_{\mathrm{PHY}}\mathcal{L}_{\mathrm{PHY}}$
            \State Update ActionDiT and Physical Head with
            $\mathcal{L}_{\mathrm{online}}$;
            update $V_\nu$ with $\mathcal{L}_{V}$
        \EndFor
    \EndFor
\EndFor
\end{algorithmic}
\end{algorithm}
\vspace{-6pt}
\section{Training and Inference}
\label{app:algorithms}

This section presents AWAM's training and inference algorithms.
Multi-intent training updates the Action and Physical Heads using FPO++ and physical supervision, with the video branch frozen and
memory fixed during collection. The proposed inference procedure reviews imagined outcomes before manipulation and updates memory from execution feedback. The Agentic Reasoning Brain remains frozen throughout; all model parameters are fixed during inference.

\subsection{World Imagination Training}
\label{app:rollout_training}
Algorithm~\ref{alg:multi_intent_training} summarizes online training. For each decision context,
the frozen Agent performs $J$ deliberation rounds and proposes
$K$ intents per round using memory and preceding execution
feedback. Each WAM intent yields one rollout of up to $H$
steps under $\pi_W^{\mathrm{old}}$ from the same simulator
snapshot. Each recorded action chunk is cached with $N$
fixed flow-time/noise probes and their old flow losses.
Memory remains fixed during collection.

GAE advantages and value targets are computed from temporally
ordered rewards, termination flags, and the old critic
$V_{\nu_{\mathrm{old}}}$. Each chunk's start-state advantage
is shared across its $N$ probes.
Eligible outcomes provide physical supervision.
With data and targets fixed, we update ActionDiT and the
Physical Head using
$\mathcal{L}_{\mathrm{online}}
=\mathcal{L}_{\mathrm{FPO++}}
+\lambda'_{\mathrm{PHY}}\mathcal{L}_{\mathrm{PHY}}$,
while training the critic separately against value targets.
The Agentic Reasoning Brain and video branch remain frozen.

\subsection{Adaptive Inference}
\label{app:adaptive_inference}

Algorithm~\ref{alg:adaptive_inference} outlines the proposed inference procedure. The Agentic Reasoning Brain uses current observations, memory, and interaction history to propose complete primitive invocations. 
For valid invocations containing manipulation channels, the World--Action Brain predicts a future $W_t$, which reviewer
$\Psi$ assesses using Physical Head predictions and fixed rules. Upon approval, the Action Head generates actions from the cached future context $f_t=F_\omega(W_t)$, and the first 28 actions are executed. Otherwise, feedback is returned to the Agent for replanning. Pure controller invocations bypass review.

Actual observations and execution feedback update the interaction history and experience memory, informing subsequent invocations. All model parameters remain frozen during inference. The episode ends when the environment terminates, the execution or planning budget is exhausted, or a valid blocked report or an infrastructure or execution error occurs. Final task outcomes are determined by the environment's native evaluation.

\begin{algorithm}[t]
\caption{Pre-execution Imagination Review and Primitive Execution}
\label{alg:adaptive_inference}
\small
\begin{algorithmic}[1]
\State \textbf{Input:} Task $g$; frozen $\pi_A,\pi_W$;
reviewer $\Psi$; memory $\mathcal{M}$; environment $\mathcal{E}$;
execution budget $H$; planning budget $B$.
\State Observe $(o,q)$; $m\leftarrow\operatorname{Read}(\mathcal{M},g)$;
$h,r\leftarrow\varnothing$; $n\leftarrow0$; $s\leftarrow\varnothing$
\While{$\mathcal{E}$ is active, execution budget remains, and $n<B$}
    \State $c\leftarrow\pi_A(g,m,o,q,h,r)$; $n\leftarrow n+1$
    \hfill{\color{gray}\textit{// Primitive Planning}}
    \State \textbf{if} $c$ is a valid blocked report
    \textbf{then} $s\leftarrow\mathrm{blocked}$; \textbf{break}
    \If{$c$ is invalid}
        \State $r\leftarrow\operatorname{InvalidFeedback}(c)$;
        append $(c,r)$ to $h$; \textbf{continue}
    \EndIf
    \If{$c$ contains manipulation channels}
        \State $z\leftarrow\operatorname{Render}(c)$;
        $(W,f)\leftarrow\operatorname{Imagine}(\pi_W,o,q,z,c)$
        \State $(d,e)\leftarrow\Psi(W,o,q,c,g)$
        \hfill{\color{gray}\textit{// Imagination Review}}
        \If{$d\neq\mathrm{pass}$}
            \State $r\leftarrow\operatorname{Feedback}(d,e)$;
            append $(c,r)$ to $h$; \textbf{continue}
        \EndIf
        \State $A\leftarrow\operatorname{Actions}(\pi_W,q,f)$;
        $\mathrm{res}\leftarrow
        \operatorname{Execute}(\mathcal{E},c,A_{1:28})$
        \hfill{\color{gray}\textit{// Primitive Execution}}
    \Else
        \State $\mathrm{res}\leftarrow
        \operatorname{Execute}(\mathcal{E},c)$
    \EndIf
    \State \textbf{if} $\mathrm{res}$ reports an infrastructure
    or execution error \textbf{then}
    $s\leftarrow\mathrm{res}$; \textbf{break}
    \State $(o,q,u)\leftarrow\operatorname{Observe}(\mathcal{E},\mathrm{res})$;
    $r\leftarrow\operatorname{Feedback}(u)$
    \State Append $(c,r)$ to $h$;
    update $\mathcal{M}$ from observed outcomes; refresh $m$
    \hfill{\color{gray}\textit{// Memory Update}}
\EndWhile
\State \Return native evaluation outcome, stop reason $s$,
and budget status
\end{algorithmic}
\end{algorithm}
\vspace{-6pt}

\section{Experience Memory}
\label{app:experience_memory}

\begin{figure*}[t]
    \centering
    \includegraphics[width=\textwidth]{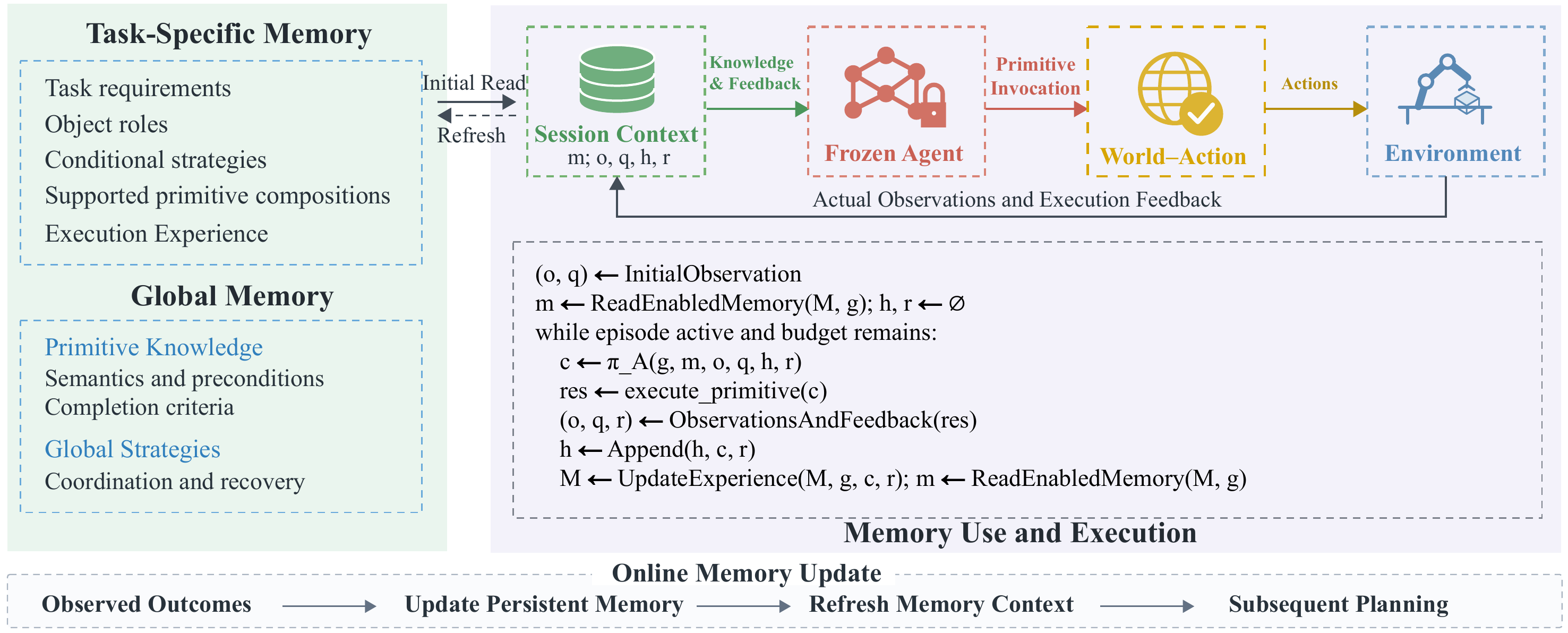}
\caption{Memory use in AWAM. Observed execution outcomes
update session history and persistent memory $\mathcal{M}$.
The refreshed memory context $m$ guides subsequent planning
and primitive selection by the frozen Agent.}
    \vspace{-12pt}
    \label{fig:experience_memory}
\end{figure*}

AWAM uses preconstructed external memory to support the frozen
Agentic Reasoning Brain. In the RoboDojo implementation, this memory
provides primitive definitions, task requirements, and conditional
methods for planning, as illustrated in
Figure~\ref{fig:experience_memory}.

\noindent\textbf{Memory Organization.}
Memory comprises global knowledge, task-family memory, and optional
task-instance experience. Global primitive knowledge defines
operation semantics, preconditions, and completion criteria.
Global strategies provide coordination and recovery principles
across tasks. Task families specify task requirements, object roles,
conditional methods, and supported primitive compositions.
The optional task-instance layer stores published experience from
actual execution episodes and is empty in the current release.

\noindent\textbf{Sources of Experience.}
The current memory release contains 43 task families: 35 draw on
3,465 training episodes from \texttt{robodojo\_primitive\_9},
and eight use independent video priors. Development episodes
are excluded from the native training priors, and video priors
are tracked separately from native demonstrations.
Global knowledge and curated strategies also retain author guidance.

\noindent\textbf{Memory Updates.}
After each execution, observed outcomes update interaction
history and experience memory online. The frozen Agent uses
the updated memory to guide subsequent primitive selection
and recovery. Predicted evidence remains distinct from
observed outcomes and is not treated as verified execution
experience.

\noindent\textbf{Retrieval and Plan Revision.}
At initialization, the system selects and validates the enabled
global files and current task family. After the initial observation,
the Agent reads this knowledge into the session context.
The memory context $m_t$ is refreshed as observed execution
outcomes are incorporated into memory. Planning uses
$x_t=(g,o_t,h_t,m_t)$ together with current feedback $r_t$.
Memory guides task interpretation and method selection, while
current observations and feedback ground primitive invocations.
Stored procedures do not establish current scene states or task
completion. The WAM receives current images, 14-dimensional
proprioception, and the complete primitive instruction.
\section{Experiments}
We systematically evaluate AWAM on public dataset, focusing on complex long-horizon tasks, compositional skill execution, and out-of-distribution generalization. Beyond task success, ablations and process diagnostics assess how dual-brain interaction, world evidence, memory, and feedback-guided replanning contribute to complex-task execution, failure recovery.

\noindent\textbf{Implementation Details.}
The Agentic Reasoning Brain uses Qwen3.5-27B in AWAM and
DeepSeek-V4.1-Flash in AWAM-Pro. Both reasoning backbones remain frozen
and are used to decompose tasks and select manipulation primitives.
In the World--Action Brain, we initialize the video DiT, text encoder,
and VAE with pretrained weights from
Wan2.2-5B~\citep{wan2025wanopenadvancedlargescale}. Given a primitive
intent, the world model predicts a future representation. A 30-layer
action Transformer with a hidden dimension of 1024 uses the resulting
future context and the current robot state to generate a 32-step action
chunk. See Appendix~\ref{app:details} for details.

\noindent\textbf{Benchmarks.}
We evaluate AWAM on two complementary benchmarks.
RoboDojo assesses generalization, memory, precision, long-horizon
execution, and open-vocabulary instruction following in simulation and the real world~\citep{chen2026robodojo}.
BiCoord tests long-horizon bimanual coordination across
subgoals~\citep{peng2026bicoord}.
Together, they cover diverse task demands, multi-stage execution, and compositional coordination.

\begin{figure*}[t]
    \centering
    \includegraphics[width=1\textwidth]{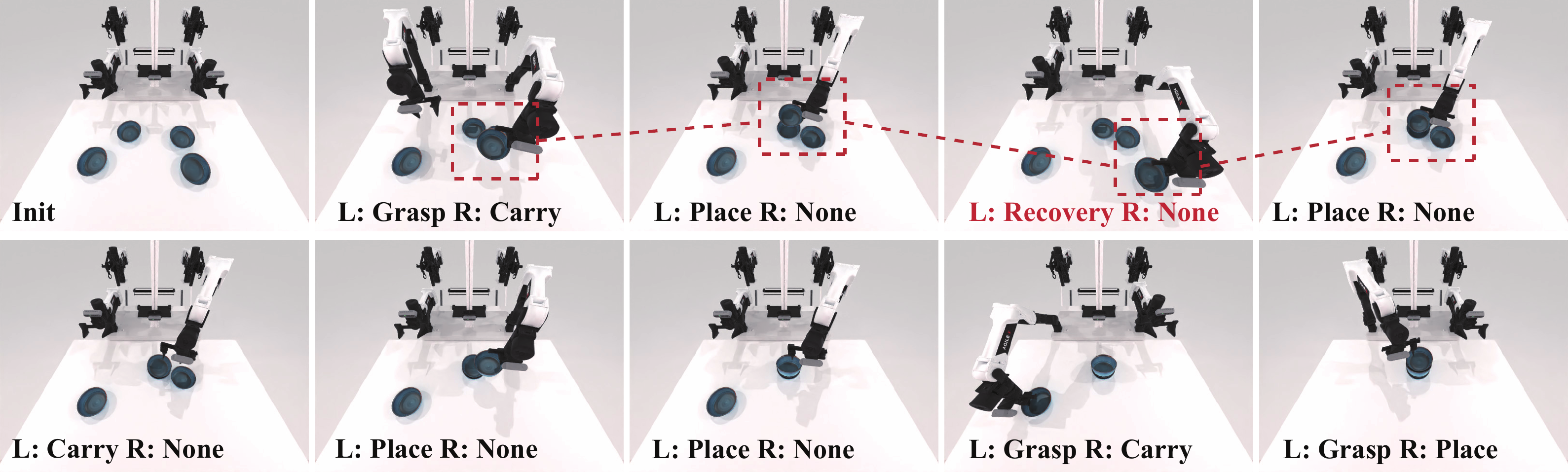}
\caption{Reasoning and failure recovery in AWAM. The agent coordinates per-arm primitives, recovers from a failed case, and resumes execution. Red annotations highlight the recovery sequence.}
    \vspace{-6pt}
    \label{fig:recovery}
\end{figure*}
\begin{table*}[t]
\centering
\caption{Performance comparison on RoboDojo across generalization, precision, long-horizon execution, memory, and open-vocabulary instruction following, measured by Score/Success Rate.}
\label{tab:robodojo_main}
\small
\resizebox{\linewidth}{!}{
\begin{tabular}{l|c|ccccc|c}
\toprule
Method
& Policy
& Generalization
& Precision
& Long-Horizon
& Memory
& Open
& Average
\\
\midrule
ACT&VLA
&0.69/0.56
&0.85/0.00
&1.73/0.92
&1.65/0.13
&0.00/0.00
&0.98/0.32
\\

$\pi_0$&VLA
&3.94/2.56
&3.56/0.75
&6.19/2.00
&3.47/2.11
&0.25/0.25
&3.48/1.53
\\

RDT-1B&VLA
&0.56/0.33
&0.38/0.00
&1.13/0.00
&0.49/0.33
&0.00/0.00
&0.51/0.13
\\

X-VLA&VLA
&10.48/6.78
&18.32/12.00
&16.53/9.75
&4.76/3.56
&0.55/0.50
&10.13/6.52
\\

$\pi_{0.5}$&VLA
&13.37/8.17
&12.40/5.50
&23.54/14.67
&5.78/4.56
&1.98/1.67
&11.41/6.91
\\

Spatial Forcing&VLA
&14.12/9.33
&17.33/10.58
&23.26/14.58
&5.43/4.11
&1.78/1.58
&12.38/8.04
\\

Hy-Embodied-0.5&VLA
&11.77/8.39
&13.81/8.00
&25.74/14.92
&13.37/12.11
&0.65/0.58
&13.07/8.80
\\

\midrule

Fast-WAM&WAM
&2.34/1.11
&1.96/0.00
&9.14/5.17
&3.55/3.44
&0.42/0.42
&3.48/2.03
\\

GigaWorld-Policy&WAM
&5.34/2.89
&6.15/1.83
&15.51/8.92
&3.46/2.22
&0.54/0.50
&6.20/3.27
\\

X-WAM&WAM
&7.39/3.33
&6.72/1.83
&17.47/9.08
&6.32/4.67
&0.57/0.25
&7.69/3.83
\\
OpenWAM&WAM
&20.71/14.83
&18.45/9.25
&34.93/25.33
&10.41/9.11
&1.41/1.08
&17.18/11.92
\\



\midrule

AWAM (Ours)
&Agent+WAM
&18.13/ 11.53
&14.21/ 6.31
&40.74/ 29.92
&25.22/ 21.69
&2.13/ 1.85
&20.08/ 14.26
\\

\bottomrule
\end{tabular}
}
\vspace{-12pt}
\end{table*}
\begin{figure*}[t]
    \centering
    \includegraphics[width=1\textwidth]{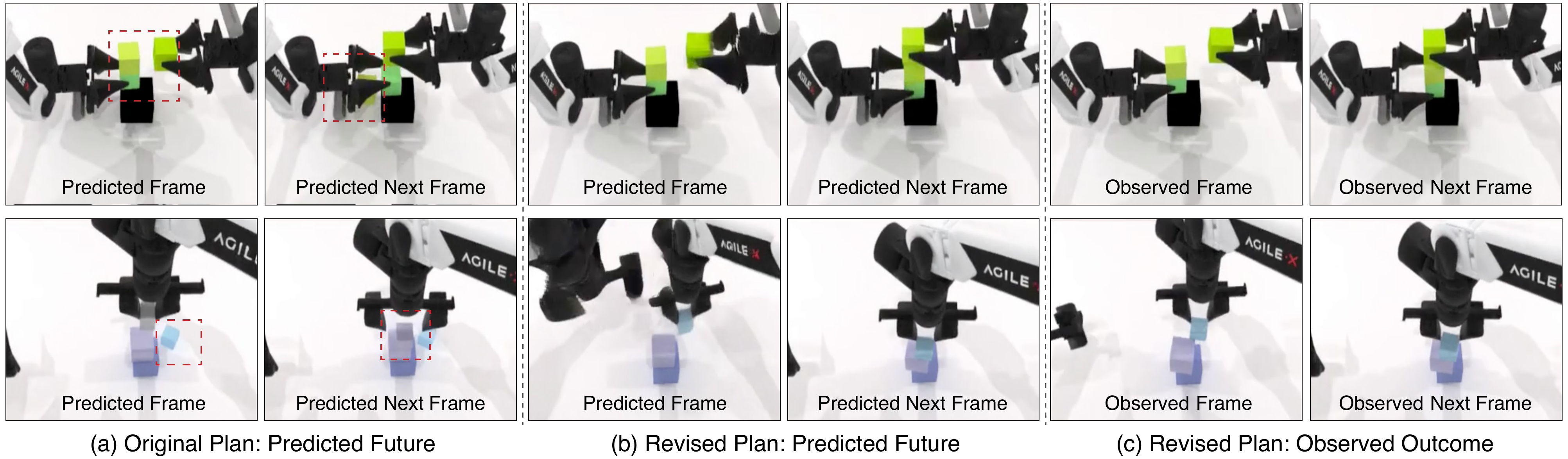}
\caption{WAM predictions before and after plan revision. Each row shows the predicted futures for (a) the original and (b) the revised primitive, followed by (c) the observed outcome after revision.}
    \vspace{-6pt}
    \label{fig:future}
\end{figure*}
\begin{table*}[t]
\centering
\small
\caption{{
Multi-task results on BiCoord. SR, SSR, and TL denote success rate, stage-wise success rate, and trajectory length. Harness+WAM uses the HarnessVLA agent with Fast-WAM in place of its original VLA policy. AWAM uses Qwen3.5 and AWAM-Pro uses DeepSeek-v4.1-flash.}}
\label{tab:bicoord_multitask}
\resizebox{1\linewidth}{!}{
\small
\setlength{\tabcolsep}{3pt}
\renewcommand{\arraystretch}{1}

\begin{tabular}{l|ccc|ccc|ccc|ccc|ccc|ccc}
\toprule

\multirow{2}{*}{Task}
& \multicolumn{3}{c|}{OpenVLA-OFT}
& \multicolumn{3}{c|}{$\pi_0$}
& \multicolumn{3}{c|}{{FastWAM}}
& \multicolumn{3}{c|}{{Harness+WAM}}
& \multicolumn{3}{c|}{AWAM}
& \multicolumn{3}{c}{{AWAM-Pro}}
\\

& SR $\uparrow$
& SSR $\uparrow$
& TL $\downarrow$
& SR $\uparrow$
& SSR $\uparrow$
& TL  $\downarrow$
& SR $\uparrow$
& SSR $\uparrow$
& TL  $\downarrow$
& SR $\uparrow$
& SSR $\uparrow$
& TL  $\downarrow$
& SR $\uparrow$
& SSR $\uparrow$
& TL  $\downarrow$
& SR $\uparrow$
& SSR $\uparrow$
& TL  $\downarrow$
\\

\midrule

Balance Roller
&49.0&74.5&137
&68.0&81.5&151
&45.0&72.5&127
&72.0 & 85.0 & 210 
&78.0& 89.0&  140
&98.0&	99.0&	131
\\

Build Bridge
&2.0&2.0&312
&44.0&47.2&289

&60.0&62.0&362
&25.0 & 32.5 & 432
&55.0& 65.0 &336
&78.0&	78.0&	269	
\\

Build Tower With Blocks
&0.0&0.0&--
&2.0&5.0&650

&21.0&25.5&558
&22.0&32.5&718
&8.0& 21.0& 700
&58.0&	68.0&	430
\\

Clean Table
&2.0&25.0&612
&6.0&46.2&658

&11.0 & 49.2 & 775
& 5.0 & 45.0 & 915
&18.0& 56.8& 756
&82.0&	94.5&	595	
\\

Collect Pens
&5.0&46.2&795
&47.0&81.8&457
&13.0&59.8&513
& 5.0 & 27.5 & 653 
&60.0& 86.0& 560
&82.0&	95.0&	572
\\

Cook
&10.0&25.0&442
&14.0&29.5&504
& 16.0 & 30.0 & 566
& 10.0 & 50.0 & 614
&28.0& 42.0& 420
&42.0&	76.5&	471
\\

Divide Block Tower
&0.0&11.1&--
&0.0&10.8&--
&  1.0 & 16.4 & 524
&  7.0 & 17.5 & 826
&8.0& 27.0& 588
&14.0&	33.8&	364
\\

Exchange Mics
& 66.0 & 67.5 & 418
& 52.0 & 59.5 & 446

& 48.0 & 48.5 & 486 
&15.0&30.0&784
&63.0& 69.0& 532
&70.0&	77.0&	500
\\

Exchange Pots
& 53.0 & 56.0 & 442
& 52.0 & 60.5 & 476
 & 40.0 & 45.0 & 537
 & 31.0 & 37.2& 826
&62.0& 67.0& 532
&26.0&	31.0&	472	
\\

Extract Bottom Block To Top
& 53.0 & 55.5 & 177
& 40.0 & 43.5 & 208
&16.0&18.0&194
& 12.0 & 14.5 & 592
&50.0& 53.5& 364
&36.0&	37.0&	257	
\\
Fetch Block With Roller
& 44.0 & 70.5 & 225
& 43.0 & 69.5 & 250
& 50.0 & 75.5 & 307
& 49.0 & 68.0 & 328
&55.0& 71.5& 252
&92.0&	96.0&	278
\\

Handover Block With Bowls
& 0.0 & 3.0 & --
& 4.0 & 9.0 & 188
&5.0 &15.5&428
&  6.0& 12.0 &   654
&12.0& 16.0& 840
&8.0&14.0& 544
\\

Jigsaw
& 0.0 & 25.2 & --
& 6.0 & 36.5 & 400
& 14.0 & 45.0 & 676
& 8.0 &36.5 & 572
&16.0& 45.5& 476
&80.0&	85.5&	452
\\

Match Blocks With Signs
& 5.0 & 8.7 & 760
& 8.0 & 18.7 & 638
&  2.0 &  7.5 &608
&  10.0 &  30.2 &   738
&17.0&	41.0&	616
&7.0&	22.0&	582

\\

Place Plate And Cup
& 55.0 & 79.2 & 315
& 25.0 & 66.2 & 302
& 52.0 & 62.8 & 374 
& 62.0 & 80.5 & 429
&40.0&	70.8&	308
&68.0&	79.0&	357
\\

Put Objects Cabinet
& 26.0 & 49.5 & 486
& 6.0 & 24.0 & 575
& 55.0 & 59.0 & 743 
& 35.0 & 62.5 & 862
&20.0& 36.0& 644
&30.0&	41.0&	564	
\\

Stack Bowls
& 1.0 & 14.0 & 400
& 12.0 & 28.0 & 471
& 28.0 & 40.0 & 540
& 50.0 & 56.7 & 648
&26.0& 39.5& 448
&58.0&	72.7&	506
\\

Sweep Block
& 44.0 & 44.0 & 234
& 61.0 & 61.0 & 253
&50.0&50.0&229

&15.0&15.0&291
& 72.0 & 72.0 & 336
&56.0&	56.0&	267
\\

\midrule

Average
& 23.1 & 36.5 & 411
& 27.2 & 43.2 & 406
&29.2&43.4&474
& 24.4 & 40.7 & 616
& 38.2 & 53.8 & 491
&54.7&	64.2&	423
\\

\bottomrule
\end{tabular}
}
\vspace{-12pt}

\end{table*}

\noindent\textbf{Baselines.}
We compare three categories of methods.
VLA policies include ACT~\citep{zhao2023learning}, RDT-1B~\citep{liu2025rdt},
OpenVLA-OFT~\citep{kim2025openvlaoft}, X-VLA~\citep{xvla},
$\pi_0$~\citep{black2024pi0}, $\pi_{0.5}$~\citep{intelligence2025pi_},
Hy-Embodied-0.5~\citep{team2026hy}, and Spatial Forcing~\citep{li2026spatial}.
WAMs for future-informed action prediction include
GigaWorld-Policy~\citep{gigaworld2026},
Fast-WAM~\citep{yuan2026fastwam}, X-WAM~\citep{x-wam},
and OpenWAM~\citep{wang2026openwam}. Agent-augmented methods are represented by HarnessVLA~\citep{zhang2026harnessvla}.


\subsection{Main Experimental Results}
\label{sec:main_results}

\noindent\textbf{RoboDojo.}
As shown in Table~\ref{tab:robodojo_main}, AWAM achieves the best overall performance, with notable gains on memory-dependent and long-horizon tasks. VLA baselines remain competitive on short-horizon precision tasks but struggle with persistent state tracking, multi-stage composition, and failure recovery. WAMs provide physical foresight through future prediction but remain constrained by finite-horizon decision making, while agent-augmented VLAs lack fine-grained physical feedback for revising high-level plans. AWAM combines experience retrieval with recurrent reason--world interaction to track progress across stages and revise intents based on predicted physical outcomes, improving long-horizon execution while preserving basic manipulation skills.

\noindent\textbf{BiCoord.}
As shown in Table~\ref{tab:bicoord_multitask},
AWAM also performs best on BiCoord, which tests long-horizon bimanual manipulation requiring precise local actions, sustained inter-arm coordination across subgoals, and dynamic role transitions. Direct action-generation methods are susceptible to coordination drift and compounding errors as the task horizon increases. AWAM instead maintains the global task structure through the
Agentic Reasoning Brain and uses the World–Action Brain to assess how candidate intents jointly affect bimanual coordination and environment dynamics before generating actions. This recurrent dual-brain interaction improves both global subtask consistency and local physical coordination, resulting in higher full-task success on tightly coupled bimanual tasks.

\noindent\textbf{Agentic Reasoning and Failure Recovery.}
As shown in Fig.~\ref{fig:recovery}, AWAM coordinates
per-arm primitives, invokes recovery after a failed placement,
and resumes task execution. This sequence illustrates its
ability to revise execution plans based on observed outcomes.
Unlike direct action-generation VLAs and WAMs without explicit
agentic reasoning, AWAM integrates high-level intent refinement
with low-level execution, enabling recovery during task execution.

\noindent\textbf{Future Predictions Before and After Plan Revision.}
Figure~\ref{fig:future} shows two block manipulation examples, with two
frames per panel. Panels (a) and (b) show the original and revised WAM
predictions, respectively, and (c) shows the observed outcome. In the
top row, the WAM initially predicts that the gripper will dislodge the
central block. The revised plan raises the gripper, leaving the block
in place. In the bottom row, the original prediction includes a
spurious block. The revised prediction removes it and correctly places
the light blue block directly above the purple block.



\subsection{Ablation Study}
\noindent\textbf{Key Component Ablation.}
The ablation results in Table~\ref{tab:component_ablation} show that task completion performance improves as the components are progressively introduced. Compared with the WAM baseline that directly maps intents to actions, adding agentic reasoning improves performance, suggesting that explicit reasoning helps organize task execution.
Incorporating experience memory yields further gains, highlighting the value of reusing past experience to support decision-making. Physical feedback provides additional benefits, suggesting that integrating physical evidence into reasoning improves the feasibility of the resulting decisions. Reinforcement learning further improves performance through interaction. Together, these results support the components’ complementary roles in high-level reasoning and low-level execution.

\begin{figure*}[t]
    \centering
    \includegraphics[width=\textwidth]{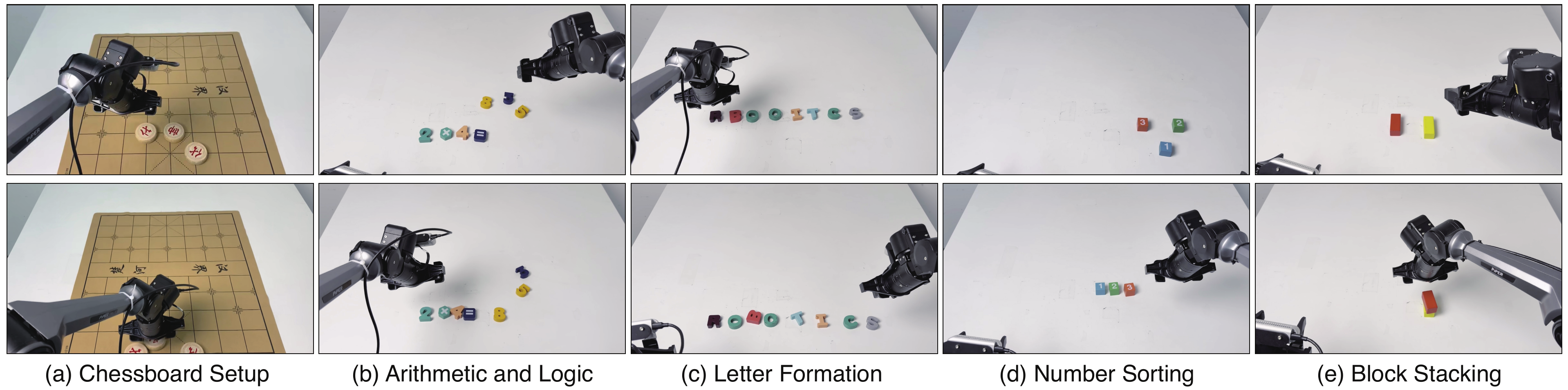}
  
\caption{Real-world tabletop manipulation with AWAM.
From left to right, the tasks are (a) chessboard setup, (b) arithmetic and logic,
(c) letter formation, (d) number sorting, and (e) block stacking.}
    \label{fig:real_data}
    \vspace{-6pt}
\end{figure*}
\begin{table*}[t]
\centering
\captionsetup{font=small,skip=2pt}
\hfill
\begin{minipage}[t]{0.48\textwidth}
\vspace{0pt}
\centering
\caption{{Component ablations on Bicoord.}
Average success rates across all categories.}
\label{tab:component_ablation}
\setlength{\tabcolsep}{12pt}

\resizebox{\linewidth}{!}{%
\begin{tabular}{lccc}
\toprule
{Model Variant}
& {SR }
& {SSR}
& {TL}
\\
\midrule
Direct Intent-to-Action &29.0 &45.0 & 420  \\
+ Agentic Reasoning &32.0 &46.5 &628  \\
+ Experience Memory &34.5 &49.0 &553  \\
+ Physical Feedback &36.0 &51.5 &504  \\
+ RL Training &38.2 &53.8 &491  \\
\bottomrule
\end{tabular}%
}
\end{minipage}
\hfill
\begin{minipage}[t]{0.46\textwidth}
\vspace{0pt}
\centering
\caption{Ablation on the number of candidate intents $K$
per training round, with $J$ and $N$ fixed.}
\label{tab:inference_efficiency}
\setlength{\tabcolsep}{12pt}
\resizebox{\linewidth}{!}{%
\begin{tabular}{cccc}
\toprule
{Candidate Intents $K$}
& {SR }
& {SSR}
& {TL}
\\
\midrule
 Times K=1&	36.0&	51.5&	448 \\
 Times K=2&	37.4&	52.8&	476 \\
 Times K=3&	38.2&	53.8&	491 \\
 Times K=4&	38.3&	53.9&	504 \\
 Times K=5&	38.5&	54.0&	532 \\

\bottomrule
\end{tabular}%
}
\end{minipage}
\hfill
\vspace{-12pt}
\end{table*}

\begin{table}[H]
    \centering
    \small
    \setlength{\tabcolsep}{8pt}
    \caption{Success rates on five real-world manipulation tasks.
Both methods use the same 30 random layouts per task. Avg.\ denotes the unweighted mean success rate across
the five tasks.}
    \label{tab:real-robot}
    \resizebox{0.98\linewidth}{!}{
    \begin{tabular}{@{}lccccc c@{}}
        \toprule
        \multirow{2}{*}{\textbf{Method}}
        & \multicolumn{5}{c}{\textbf{Real-World Tasks}}
        & \multirow{2}{*}{\textbf{Avg.\ (\%)}} \\
        \cmidrule(lr){2-6}
        & \shortstack{Block\\Stacking (\%)}
        & \shortstack{Number\\Sorting (\%)}
        & \shortstack{Arithmetic\\and Logic (\%)}
        & \shortstack{Chessboard\\Setup (\%)}
        & \shortstack{Letter\\Formation (\%)}
        & \\
        \midrule
        $\pi_{0.5}$ & (22/30) 73.3 & (18/30) 60.0   & (13/30) 43.3 & (21/30) 70.0 & (18/30) 60.0 & 61.3 \\
        AWAM (Ours) & (25/30) 83.3 & (23/30) 76.7 & (26/30) 86.7 & (25/30) 83.3 & (23/30) 76.7 & 81.3 \\
        \bottomrule
    \end{tabular}
    }
    \vspace{-12pt}
\end{table}

\noindent\textbf{Candidate Intent Count.}
Table~\ref{tab:inference_efficiency} varies $K$ while keeping $J$ and $N$ fixed.
Task and stage-wise success rates increase with $K$, with
diminishing gains beyond $K=3$. Trajectory length also
increases, indicating that more candidate rollouts during
training do not necessarily yield shorter execution trajectories.
With $J$ fixed, increasing $K$ also increases the amount
of collected execution data.

\subsection{Real-World Experiments}
\label{real_robot}
We evaluate AWAM on five tabletop tasks using the Piper X robot from
AgileX Robotics (Fig.~\ref{fig:real_data}): Chessboard Setup, Arithmetic
and Logic, Letter Formation, Number Sorting, and Block Stacking.
AWAM uses a cloud-hosted LLM for planning and a local World--Action Brain
for action generation, with peak GPU memory usage of 23.5~GB on a single
NVIDIA RTX~4090.
We compare AWAM with $\pi_{0.5}$~\citep{intelligence2025pi_} on the same
30 randomized layouts per task, with fixed model
parameters and identical robot hardware, observations, instructions, and
execution budgets. Table~\ref{tab:real-robot} reports per-task
success rates and their unweighted mean.

\section{Conclusion}
\label{sec:conclusion}
We introduced AWAM, a dual-brain framework that couples agentic
reasoning with world modeling through recurrent interaction.
The Agentic Reasoning Brain generates structured intents,
while the World--Action Brain predicts their consequences,
decodes actions from world representations, and provides
physical evidence for intent refinement.
With the reasoning backbone frozen, AWAM integrates experience
memory with world--action learning and reinforcement learning
to support single-path execution with adaptive replanning.
Experiments on RoboDojo and BiCoord demonstrate improvements
in task completion, long-horizon execution, and bimanual
coordination. Ablations support the complementary contributions
of agentic reasoning, experience memory, physical feedback,
and reinforcement learning.
Future work will explore open-world tasks, cross-embodiment
transfer, and continual learning from physical experience.

\newpage
\printbibliography[title={References}]

\newpage
\appendix

\section{Implementation Details}
\label{app:details}
\subsection{Dual-Brain Architecture and Interfaces}
\label{app:dual_brain_architecture}
AWAM integrates a frozen reasoning backbone with trainable world and action modules through structured intents, physical feedback, and experience memory.
Sparse world--action interaction and cached future context support efficient action generation, while fixed physical review rules guide intent acceptance and adaptive replanning. Table~\ref{tab:awam_configuration} summarizes the architecture, training configuration, and inference settings.
\begin{table*}[h]
\centering
\caption{Architecture and implementation settings of AWAM.}
\label{tab:awam_configuration}
\small
\setlength{\tabcolsep}{6pt}
\renewcommand{\arraystretch}{1}
\resizebox{0.9\textwidth}{!}{
\begin{tabular}{l|l|l}
\toprule
\textbf{Category} & \textbf{Configuration} & \textbf{Setting} \\
\midrule

\multirow{3}{*}{Reasoning Brain}
& Reasoning backbone & Qwen3.5-27B \\
& Backbone parameters & Frozen \\
& Experience memory & Updated separately \\
\midrule

\multirow{8}{*}{World and Action Modules}
& Pretrained backbone & Wan2.2-5B \\
& Initialized components & Video DiT, text encoder, and VAE \\
& Multi-camera input & Spatially assembled images \\
& Visual sequence length & 9 frames \\
& Temporal sampling stride & 4 \\
& Action chunk length & 32 \\
& Action Transformer depth & 30 layers \\
& Action hidden dimension & 1024 \\
\midrule

\multirow{5}{*}{World--Action Interaction}
& Interaction stages & 8 \\
& Interaction interval & 4 stages \\
& Feature aggregation & Interval KV-Fusion \\
& Future-context reuse & Cached throughout each action chunk \\
& Action conditioning & Future context and robot state \\
\midrule

\multirow{7}{*}{Training}
& World/action training objective & Flow matching \\
& Optimizer & AdamW \\
& Initial learning rate & $10^{-4}$ \\
& Weight decay & $0.01$ \\
& Learning-rate schedule & Cosine decay \\
& Training precision & BF16 \\
& Maximum gradient norm & $1.0$ \\
\midrule

\multirow{2}{*}{Action Inference}
& Numerical integration steps & 10 \\
& Guidance scale & $1.0$ \\
\bottomrule
\end{tabular}
}
\end{table*}

\noindent\textbf{Agentic Reasoning Brain Details.}
The Agentic Reasoning Brain $\pi_A$ uses Qwen3.5 with frozen parameters. It combines the task instruction and current observations with interaction history and retrieved experience to plan the subtask sequence and specify the primitive invocation for the next operation. These outputs form the intent representation supplied to the World Brain. Predicted physical evidence, review decisions, and observed
execution outcomes are recorded in the session context and run records, allowing subsequent decisions to use physical predictions alongside actual feedback. The architecture does not require a particular reasoning backbone. Other compatible large language models can replace Qwen3.5 using the same input and structured-output interfaces.

\noindent\textbf{World--Modeling Brain Details.}
The video DiT, text encoder, and VAE in $\pi_W$ are initialized from Wan2.2-5B~\citep{wan2025wanopenadvancedlargescale}. Multi-camera images are combined into a shared spatial layout before VAE encoding. Video supervision uses nine frames sampled at a temporal stride of four. The model predicts 32 action steps per chunk and, by default, executes the first 28 steps before acquiring new observations. The action module uses a 30-layer Transformer with a hidden dimension of 1024. Intent information conditions world modeling, while action generation is conditioned on the resulting future context and the current robot state.

\noindent\textbf{World--Action Interaction Details.}
World features are injected at eight selected stages of the 30-layer action network, with an interaction interval of four. Interval KV-Fusion~\citep{zhao2026faster} aggregates video features over the corresponding depth intervals, while the remaining action
layers update only action features.
The future context is computed once and cached throughout the generation of each action chunk. World evidence informs a fixed review rule: approved intents
proceed to action generation, while rejected or uncertain
proposals return feedback to $\pi_A$ for intent refinement
and replanning.

\noindent\textbf{Optimization and Inference.}
During supervised fine-tuning, visual and action generation are trained using flow matching. We optimize the trainable components with AdamW and a weight decay of 0.01. The current SFT configurations specify initial learning rates of $10^{-4}$.
The learning rate follows a cosine decay schedule. Training uses BF16 precision and gradient norm clipping with a maximum norm of $1.0$. The reasoning backbone remains frozen, while experience memory is maintained separately. At inference time, actions are sampled using 10 numerical integration steps with a guidance scale of $1.0$.

\subsection{Data Construction and Details}

\subsubsection{Data Construction}
\label{app:data_pipeline}
AWAM's dual-brain architecture requires data that support task-level
reasoning, Primitive-level manipulation, and physical feedback modeling.
We therefore organize robot demonstrations into a Task--Skill--Primitive
hierarchy linked to action sequences, derive Primitive-conditioned
action supervision, and obtain physical outcome labels through
candidate execution. 
\begin{figure*}[t]
    \centering
    \includegraphics[width=0.95\textwidth]{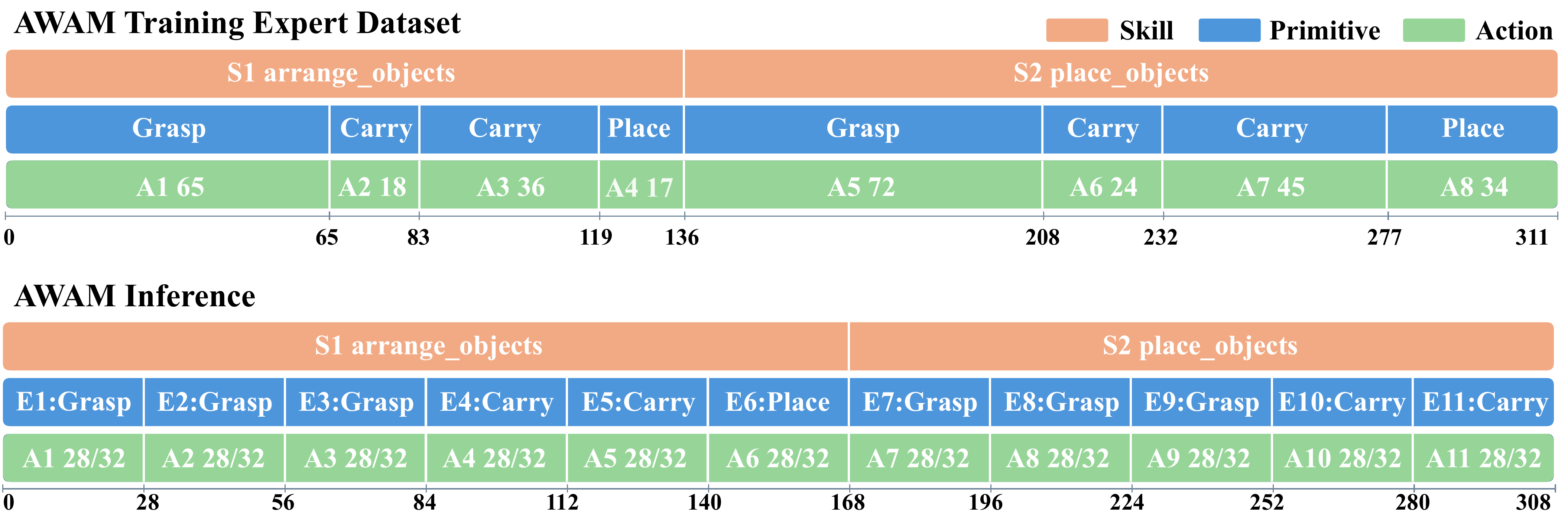}
\caption{{Expert and AWAM execution timelines on the BiCoord bridge task.}
Rows show Skill, Primitive, and Action segments. AWAM executes 28 steps per 32-step chunk. Timelines are independently normalized, without frame-wise alignment.}
    \label{fig:primitive}
\end{figure*}

\paragraph{Data Sources and Annotation.}
We use demonstrations from BiCoord and RoboDojo, together with available replay records, and apply the same hierarchical annotation procedure to real-world manipulation trajectories. We first align visual observations, robot states, actions, and physical evidence. Primitive boundaries are identified from changes in motion and gripper state and verified using semantic and physical evidence. For bimanual trajectories, merging left- and right-arm boundaries yields composition units, each containing at most one Primitive per arm and specifying grounded object roles, inter-arm dependencies, and termination conditions. Primitive sequences are then grouped into Skills according to local goals while preserving their task-level execution order, as shown in Fig~\ref{fig:primitive}. Shared rules convert each object-grounded Primitive step into a conditioning instruction during both data construction and runtime. Trajectories with incomplete annotations are excluded. 



\paragraph{Interactive Action Primitive Annotation Tool.}
We develop an interactive tool for the structured annotation of action
primitives in bimanual manipulation trajectories. The tool displays key
events, including gripper state changes and pauses in arm motion, on
separate timelines for the left and right arms. Observation frames and
key-frame thumbnails help annotators interpret the manipulation process.
Annotators can specify skill names, execution order, and dependencies,
as well as action primitive sequences for each arm. The annotations can
be exported in a structured format to support subsequent data processing
and model training.

\subsection{Data Formats and Examples}
\paragraph{Hierarchical Trajectory Representation.}
Figure~\ref{fig:primitive} illustrates the resulting
Skill--Primitive--Action hierarchy using an expert demonstration
and an AWAM execution. Each trajectory associates Skills with
Primitive segments and their corresponding action intervals.
In the illustrated AWAM execution, each prediction window contains
32 action steps, of which the first 28 are executed before acquiring
a new observation for the next prediction. The records retain
repeated Primitive invocations and actual execution intervals
for feedback analysis and memory construction.

\noindent\textbf{Primitive Action Sample.}
Figure~\ref{fig:primitive_action_sample} shows a training sample for the \texttt{balance\_roller}
task in \texttt{primitivewam\_13}. The sample specifies a coupled
bimanual grasp of the roller, with the complete conditioning
instruction ``Use both arms to pick up the roller.''
The current frame index is 0, and the composition interval is
$[0,76]$. Action supervision consists of 14-D native joint-action
vectors at indices 1--32. All 32 time steps have valid supervision,
so no temporal padding is required.
\begin{figure}[t]
\centering
\begin{minipage}{0.97\linewidth}
\begin{lstlisting}[style=awamjson]
{
  "sample_id": "bicoord__balance_roller__clean__0__w0",
  "schema": "primitivewam-v12-dataset-record",
  "schema_version": "12.0",
  "split": "train",
  "frame_index": 0,
  "composition_interval": [0, 76],
  "wam_condition": "Use both arms to pick up the roller.",
  "step": {
    "schema_version": "11.0",
    "dependency": "coupled",
    "left": {
      "controller": "left",
      "primitive": "Grasp",
      "target": "the roller"
    },
    "right": {
      "controller": "right",
      "primitive": "Grasp",
      "target": "the roller"
    }
  },
  "target": {
    "action_ref": {
      "dataset": "/joint_action/vector",
      "index_semantics": "native_joint_action[t+1]",
      "start": 1,
      "valid_length": 32,
      "horizon": 32,
      "dimension": 14
    }
  }
}
\end{lstlisting}
\end{minipage}
\caption{A primitive-conditioned training sample for the BiCoord
\texttt{balance\_roller} task, specifying a coupled
bimanual grasp. Given the observation at frame 0, action supervision
comprises 32 consecutive 14-dimensional native joint-action vectors
at indices 1--32. Image and proprioceptive references, selected
provenance fields, and validity masks are omitted. Task and provenance
metadata are excluded from WAM inputs.}
\label{fig:primitive_action_sample}
\end{figure}

\noindent\textbf{Physical Feedback Sample.}
Figure~\ref{fig:physical_feedback_sample} shows an offline Physical Head supervision record
derived from demonstration annotations and bound to the instruction ``Grasp the number 6 with the left arm.'' Its proxy outcome label is \texttt{intermediate}.
The \texttt{success}, \texttt{intermediate}, and \texttt{failure} labels participate in supervision, while \texttt{forbidden\_collision} and \texttt{interaction\_formed} are excluded from the loss. The mask indicates training eligibility at the current stage, not label certainty or physical verification. This record does not constitute evidence of online model execution, and the masked zero entries do not establish the absence of forbidden collisions or interaction formation.
\begin{figure}[h]
\centering
\begin{minipage}{0.97\linewidth}
\begin{lstlisting}[style=awamjson]
{
  "schema": "physical-head-supervision/v1",
  "sample_id": "robodojo__arrange_largest_number__arx_x5__0__f0",
  "label_order": [
    "success",
    "intermediate",
    "failure",
    "forbidden_collision",
    "interaction_formed"
  ],
  "target": [
    0,
    1,
    0,
    0,
    0
  ],
  "mask": [
    true,
    true,
    true,
    false,
    false
  ],
  "verification": "ANNOTATION_DERIVED",
  "mask_semantics": "eligible for training loss at the current stage; not label certainty",
  "binding": {
    "instruction": "Grasp the number 6 with the left arm."
  }
}
\end{lstlisting}
\end{minipage}
\caption{An offline Physical Head supervision sample derived from RoboDojo annotations. The proxy outcome label is \texttt{intermediate}. Only the three outcome labels contribute to the training loss. The two event labels are masked out. The mask
specifies supervision eligibility, not label certainty.}
\label{fig:physical_feedback_sample}
\end{figure}

\noindent\textbf{Memory Sample.}
Figure~\ref{fig:strategy_memory_sample} shows a structured memory sample for the
\texttt{stack\_bowls\_two} task in RoboDojo.
The record links a bowl-stacking skill to a left-arm
\texttt{Lift} primitive, with a bowl as the target and
the table as the reference.
Its sequence context places \texttt{Lift} after
\texttt{Grasp} and before \texttt{Carry}, supporting
primitive selection and sequencing.
This knowledge does not establish current scene states
or indicate that an action has been completed.
The initial memory release contains no task-instance
experience records. Global strategies are stored in
global memory files, while task-level methods are
maintained within task families; no separate
Strategy Memory layer is defined.
\begin{figure}[t]
\centering
\begin{minipage}{0.97\linewidth}
\begin{lstlisting}[style=awamjson]
{
  "task": "stack_bowls_two",
  "task_family": "stacking",
  "skill_instruction": "place the selected bowl on the reference bowl",
  "skill_relations": ["placed_at"],
  "dependency": "none",
  "primitive_pattern": {
    "left": {
      "primitive": "Lift",
      "target_category": "bowl",
      "reference_category": "table"
    },
    "sequence_context": {
      "previous_primitives": ["Grasp"],
      "next_primitives": ["Carry"],
      "skill_position": 1,
      "skill_length": 4
    }
  },
  "source_split": "train"
}
\end{lstlisting}
\end{minipage}
\caption{A structured memory sample for the
\texttt{stack\_bowls\_two} task (selected fields). The record links
a bowl-stacking skill to a left-arm \texttt{Lift} primitive, including object categories and its sequential context between \texttt{Grasp} and \texttt{Carry}. It represents reusable procedural knowledge, not the current scene state or completed actions.}
\label{fig:strategy_memory_sample}
\end{figure}

\subsection{Dataset Statistics}
\subsubsection{Hierarchical Decomposition and Statistics of BiCoord}
\label{app:bicoord_statistics}

To support the dual-brain paradigm, we segment BiCoord manipulation trajectories into skill-level and primitive-level units, yielding BiCoord-Skill and BiCoord-Primitive, respectively. BiCoord-Skill enriches the Agentic Reasoning Brain's memory with knowledge of the skills required to complete a task, while BiCoord-Primitive provides elementary manipulation units for training
the World--Action Brain.

\begin{figure*}[h]
    \centering
    \includegraphics[width=1\textwidth]{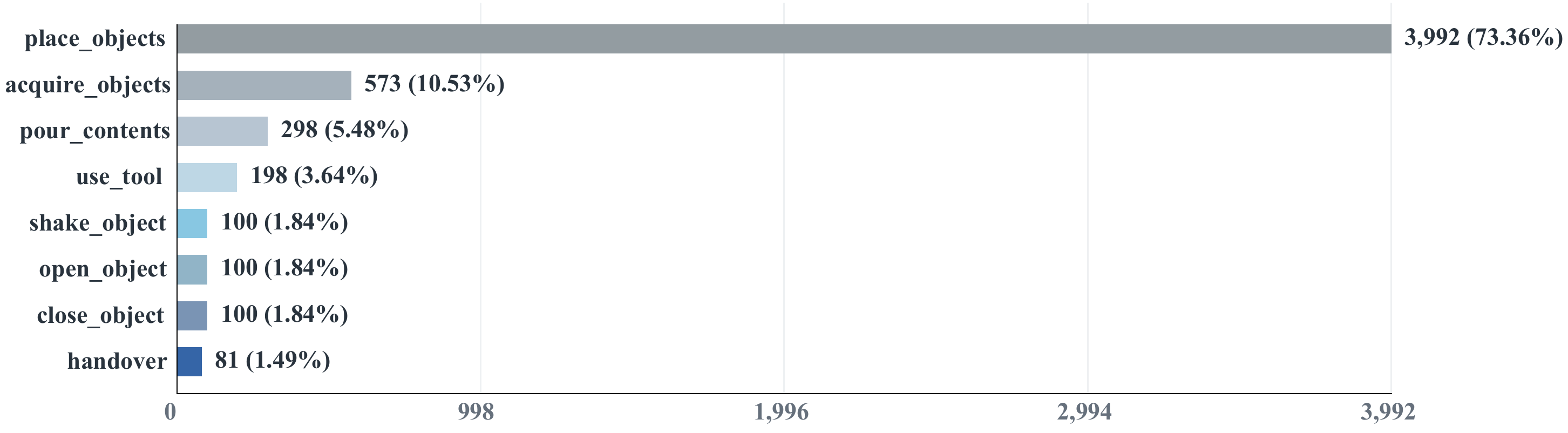}
    
\caption{Skill distribution in BiCoord-Skill across 8 annotation categories. Bar lengths indicate annotation counts, with corresponding percentages shown alongside.}
    \label{fig:skill_bicoord}
\end{figure*}
\noindent\textbf{BiCoord Skill.}
As shown in Fig.~\ref{fig:skill_bicoord}, BiCoord-Skill contains
5,442 annotations across 8 skill categories. Object placement accounts for 73.36\% of the annotations, followed by object acquisition (10.53\%), pouring (5.48\%), and tool use (3.64\%). Shaking, opening, and closing each account for 1.84\%, while
handover contributes 1.49\%. Organized around complete skills, this representation supports the Agentic Reasoning Brain in
tracking task progress and selecting subsequent operations.
\begin{figure*}[h]
    \centering
    \includegraphics[width=0.9\textwidth]{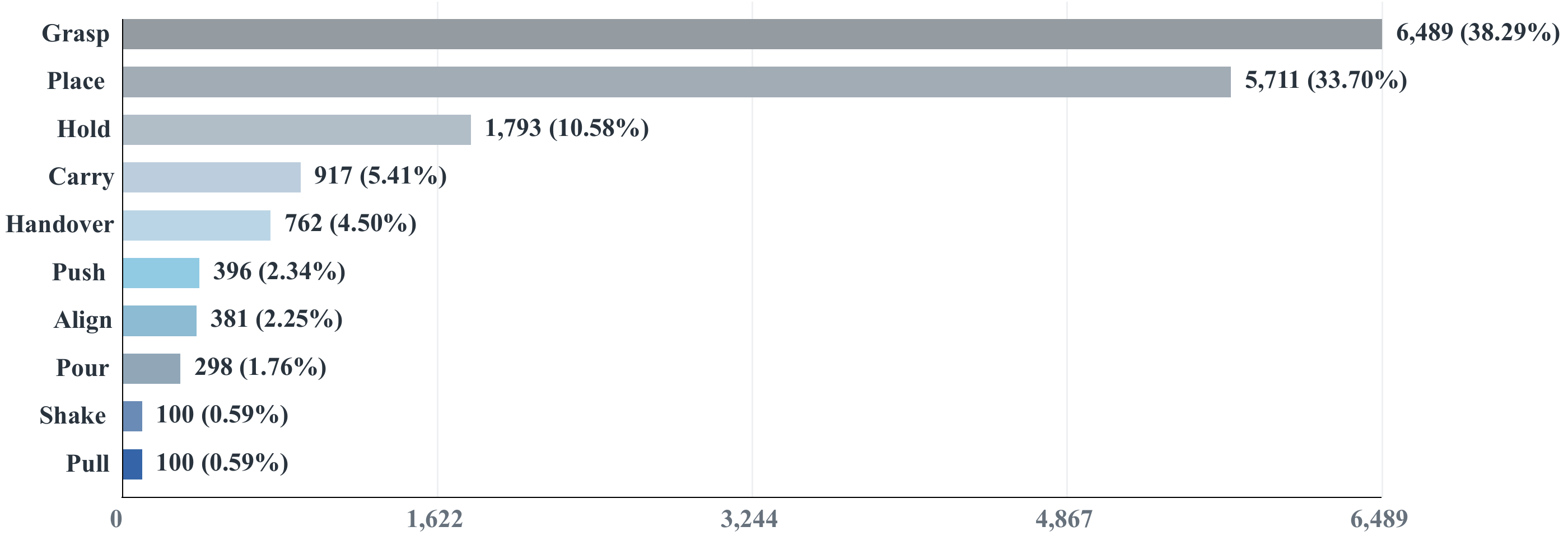}
    
\caption{Primitive distribution in BiCoord-Primitive across
10 primitive categories. Bar lengths indicate annotation counts, with corresponding percentages shown alongside.}
    \label{fig:primitive_bicoord}
\end{figure*}

\noindent\textbf{BiCoord Primitive.}
As shown in Fig.~\ref{fig:primitive_bicoord}, BiCoord-Primitive
contains 16,947 annotations across 10 primitive categories. Grasping and placing account for 38.29\% and 33.70\% of the
annotations, respectively, followed by holding (10.58\%),
carrying (5.41\%), and handover (4.50\%). Pushing, alignment, and pouring contribute 2.34\%, 2.25\%, and 1.76\%, respectively, while shaking and pulling each account for 0.59\%. This representation supports physical interaction modeling and primitive-conditioned action generation in the World--Action Brain.

\subsubsection{Hierarchical Decomposition and Statistics of RoboDojo}
\label{app:robodojo_statistics}

Following the same hierarchical decomposition, we segment RoboDojo trajectories into skill-level and primitive-level units, yielding RoboDojo-Skill and RoboDojo-Primitive, respectively, to support high-level task reasoning and low-level action execution.

\begin{figure*}[h]
    \centering
    \includegraphics[width=1\textwidth]{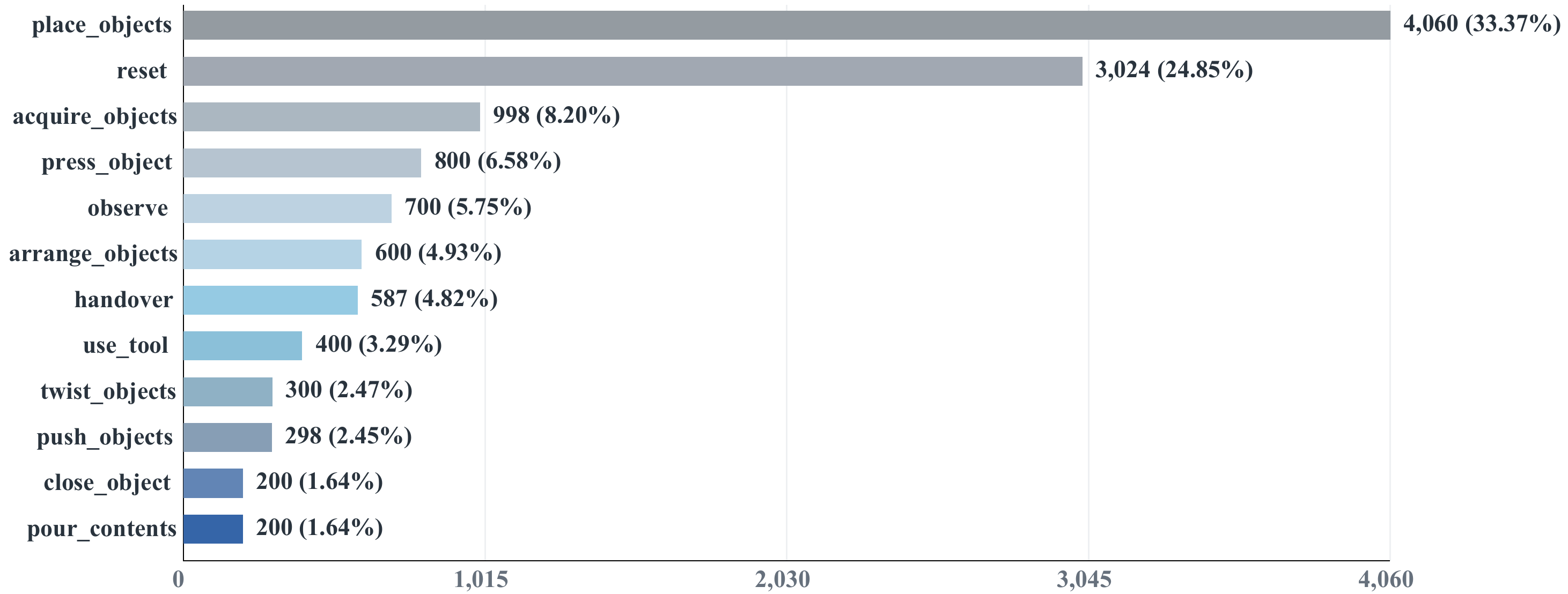}
    
\caption{Skill distribution in RoboDojo-Skill across 12 annotation categories. Bar lengths indicate annotation counts, with corresponding percentages shown alongside.}
    \label{fig:skill_robodojo}
\end{figure*}
\noindent\textbf{RoboDojo Skill.}
As shown in Fig.~\ref{fig:skill_robodojo}, RoboDojo-Skill contains
12,167 annotations across 12 categories, including the control
categories \texttt{reset} and \texttt{observe}. Object placement
and reset account for 33.37\% and 24.85\% of the annotations,
respectively. Object acquisition, pressing, observation, and
object arrangement contribute 8.20\%, 6.58\%, 5.75\%, and 4.93\%,
respectively. Handover, tool use, twisting, and pushing account
for 4.82\%, 3.29\%, 2.47\%, and 2.45\%, while closing and pouring
each contribute 1.64\%. This representation organizes trajectories
into manipulation and control stages, supporting task-progress
tracking and subsequent operation selection.

\begin{figure*}[h]
    \centering
    \includegraphics[width=\textwidth]{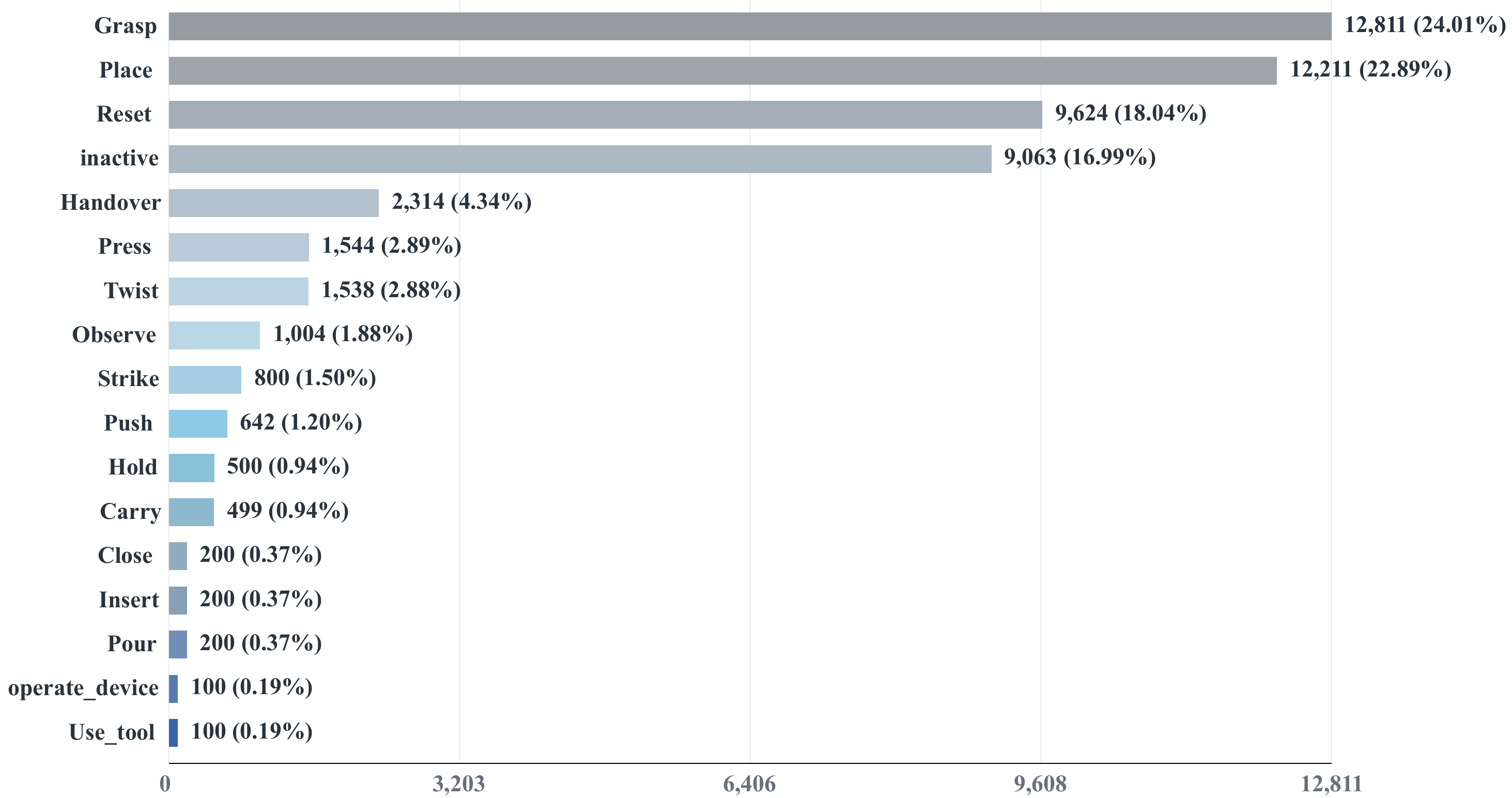}
    
\caption{Primitive distribution in RoboDojo-Primitive across 17 action, control, and state categories. Bar lengths indicate annotation counts, with corresponding percentages shown alongside.}
    \label{fig:primitive_robodojo}
\end{figure*}
\noindent\textbf{RoboDojo Primitive.}
As shown in Fig.~\ref{fig:primitive_robodojo}, RoboDojo-Primitive
contains 53,350 annotations across 17 action, control, and state
categories. Grasping, placing, reset, and inactivity account for
24.01\%, 22.89\%, 18.04\%, and 16.99\% of the annotations,
respectively. Handover, pressing, twisting, and observation
contribute 4.34\%, 2.89\%, 2.88\%, and 1.88\%, respectively,
while each remaining category accounts for at most 1.50\%.
This representation captures manipulation primitives alongside
controller operations and inactive states; manipulation annotations
support primitive-conditioned action generation in the
World--Action Brain.

\section{Theoretical Analysis}
\label{app:theoretical_analysis}

\subsection{Proof of Proposition 1}
\label{app:proof_proposition}
We fix a decision step $t$, a deliberation round $j$, and
a finite candidate set $\mathcal{C}_t^j$.
All distributions below are conditioned on the current
reasoning context and candidate set.
These conditions are omitted for brevity. Let $\pi_A^j$ be the normalized candidate prior and define
its support as
$\mathcal{S}_t^j=
\{z\in\mathcal{C}_t^j:\pi_A^j(z)>0\}$.
Assume that the estimated execution-success likelihood
$\ell_W^j(z)$ satisfies $0<\ell_W^j(z)\leq1$
for every $z\in\mathcal{S}_t^j$.
The normalizing constant and reference posterior are
\begin{equation}
\label{eq:app_reference_posterior}
Z_j
=
\sum_{z\in\mathcal{S}_t^j}
\pi_A^j(z)\ell_W^j(z),
\qquad
q_j^*(z)
=
\frac{\pi_A^j(z)\ell_W^j(z)}{Z_j}.
\end{equation}
The assumptions imply $0<Z_j\leq1$.
We set $q_j^*(z)=0$ outside $\mathcal{S}_t^j$.

For any candidate distribution $q$ satisfying $q\ll\pi_A^j$,
consider the objective
\begin{equation}
\label{eq:app_variational_functional}
\mathcal{F}_j(q)
=
\mathbb{E}_{z\sim q}[\log\ell_W^j(z)]
-
D_{\mathrm{KL}}(q\Vert\pi_A^j).
\end{equation}
Substituting the definition of $q_j^*$ gives
\begin{equation}
\label{eq:app_kl_decomposition}
\begin{aligned}
D_{\mathrm{KL}}(q\Vert q_j^*)
&=
\sum_{z\in\mathcal{S}_t^j}
q(z)\log
\frac{q(z)Z_j}{\pi_A^j(z)\ell_W^j(z)}\\
&=
D_{\mathrm{KL}}(q\Vert\pi_A^j)
-
\mathbb{E}_{z\sim q}[\log\ell_W^j(z)]
+
\log Z_j.
\end{aligned}
\end{equation}
Therefore,
\begin{equation}
\label{eq:app_optimality_gap}
\mathcal{F}_j(q)
=
\log Z_j-D_{\mathrm{KL}}(q\Vert q_j^*).
\end{equation}
By nonnegativity of the KL divergence,
$\mathcal{F}_j(q)\leq\log Z_j$,
with equality if and only if $q=q_j^*$.
Hence, $q_j^*=\pi_{\mathrm{RW}}^j$ is the unique optimizer
of Eq.~\ref{eq:reason_world_variational}.
Moreover, the objective gap is exactly
$\mathcal{F}_j(q_j^*)-\mathcal{F}_j(q)
=D_{\mathrm{KL}}(q\Vert q_j^*)$,
which completes the proof.

\subsection{Assumptions and Scope}
\label{app:theory_scope}

The proposition applies to a fixed finite candidate set with a normalized prior and positive execution-success likelihoods on the prior's support. It characterizes the unique variational optimizer for the given prior and likelihood estimates. The likelihood $p_W(y=1\mid o_t,z)$ denotes the model-estimated probability of intent success and should condition on all information relevant to that outcome. The posterior serves as a theoretical reference.
Implemented inference uses Physical Head predictions and fixed review rules to approve an invocation or request replanning, without explicitly computing this posterior. The proposition therefore does not guarantee optimal review decisions, convergence across rounds, or monotonic improvement
in task success. Practical gains depend on predictive accuracy and are assessed through execution-level experiments.

\begin{figure*}[t]
    \centering
    \includegraphics[width=0.9\textwidth]{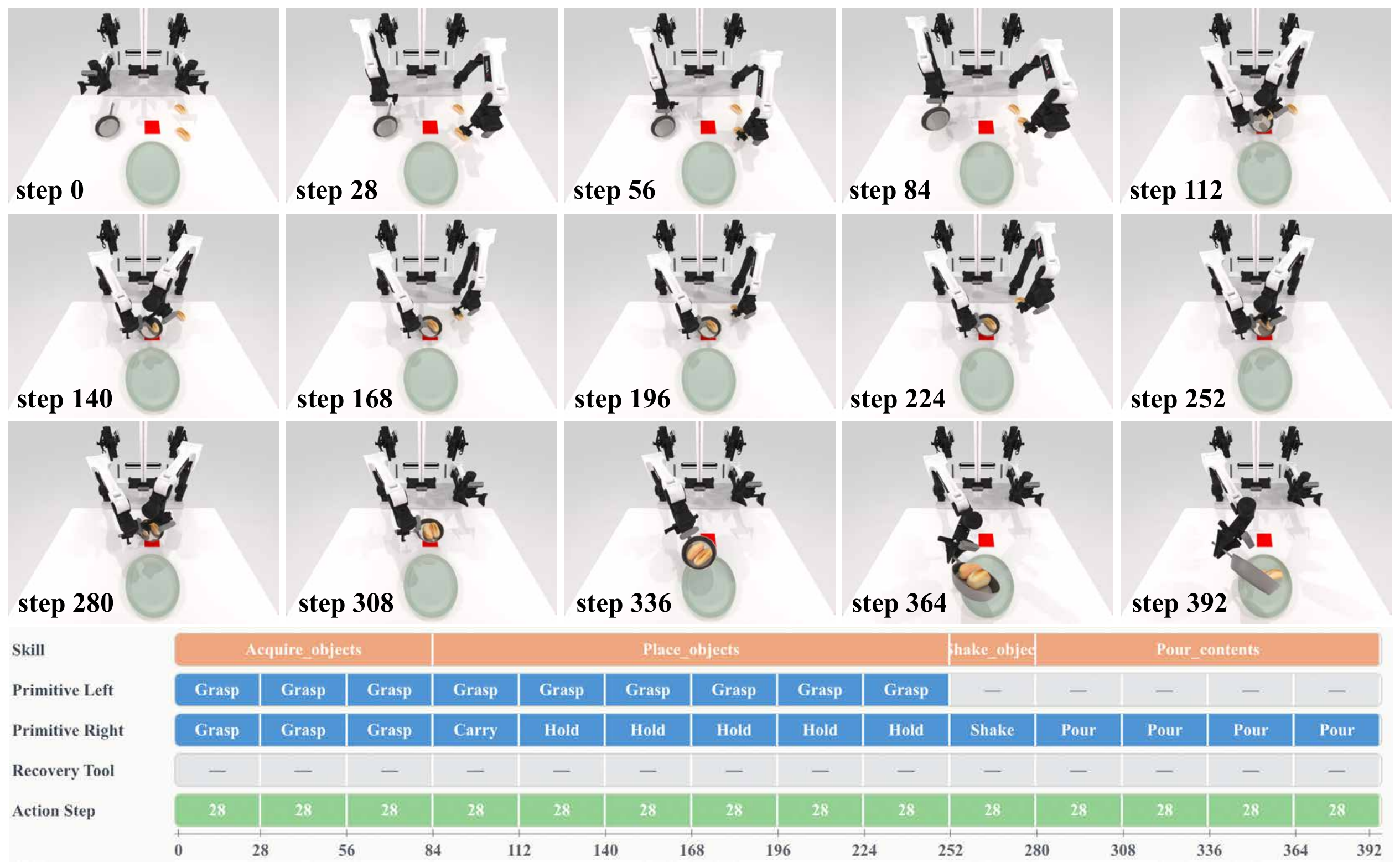}
    \caption{Bimanual cooking. AWAM coordinates food placement, pan
    shaking, and pouring. The timeline shows the per-arm primitives
    used at each stage.}
    \label{fig:bimanual_cooking}
\end{figure*}
\section{Benchmarks and Evaluation}
\label{app:benchmarks_evaluation}
\subsection{RoboDojo}
RoboDojo is a unified benchmark for generalist robotic manipulation, comprising 42 simulation tasks and 18 real-world tasks~\citep{chen2026robodojo}.
Its simulation suite uses the ARX X5 bimanual platform in Isaac Sim and supports parallel evaluation across diverse tasks. The accompanying simulation dataset contains 3,500 demonstration trajectories covering 34 training tasks and an auxiliary augmentation set, while the eight Open tasks are reserved for evaluating transfer to unseen task specifications.

Our evaluation covers generalization, precision, long-horizon execution, memory, and open-vocabulary instruction following. These categories assess robustness to environmental changes,
fine-grained control, multi-stage completion, retention of historical observations, and skill recombination under novel instructions.
Table~\ref{tab:robodojo_main} reports score / success rate for each category, measuring partial task progress and
complete task success, respectively.
The overall results are computed by averaging the corresponding metrics across the five categories.

\subsection{BiCoord}
BiCoord is built on RoboTwin 2.0~\citep{mu2025robotwin2} and comprises 18 bimanual manipulation tasks requiring sustained inter-arm coordination and dynamic role exchange across multiple subgoals~\citep{peng2026bicoord}. Each task provides 100 successful demonstration trajectories with stage-level annotations of temporal boundaries, subgoals, and arm behaviors. These annotations support evaluation of intermediate progress alongside complete task execution.

We use the multi-task learning setting and report results in Table~\ref{tab:bicoord_multitask}.
The benchmark protocol evaluates each task over 100 rollouts using Success Rate (SR), Stage-Wise Success Rate (SSR),
and Trajectory Length (TL). SR measures full-task completion, while SSR measures
partial progress through a weighted sum of completed stages. TL counts execution steps in successful trajectories. When computing average TL, tasks on which any compared method has zero success are excluded to ensure a common comparison set.



\begin{figure*}[t]
    \centering
    \includegraphics[width=0.9\textwidth]{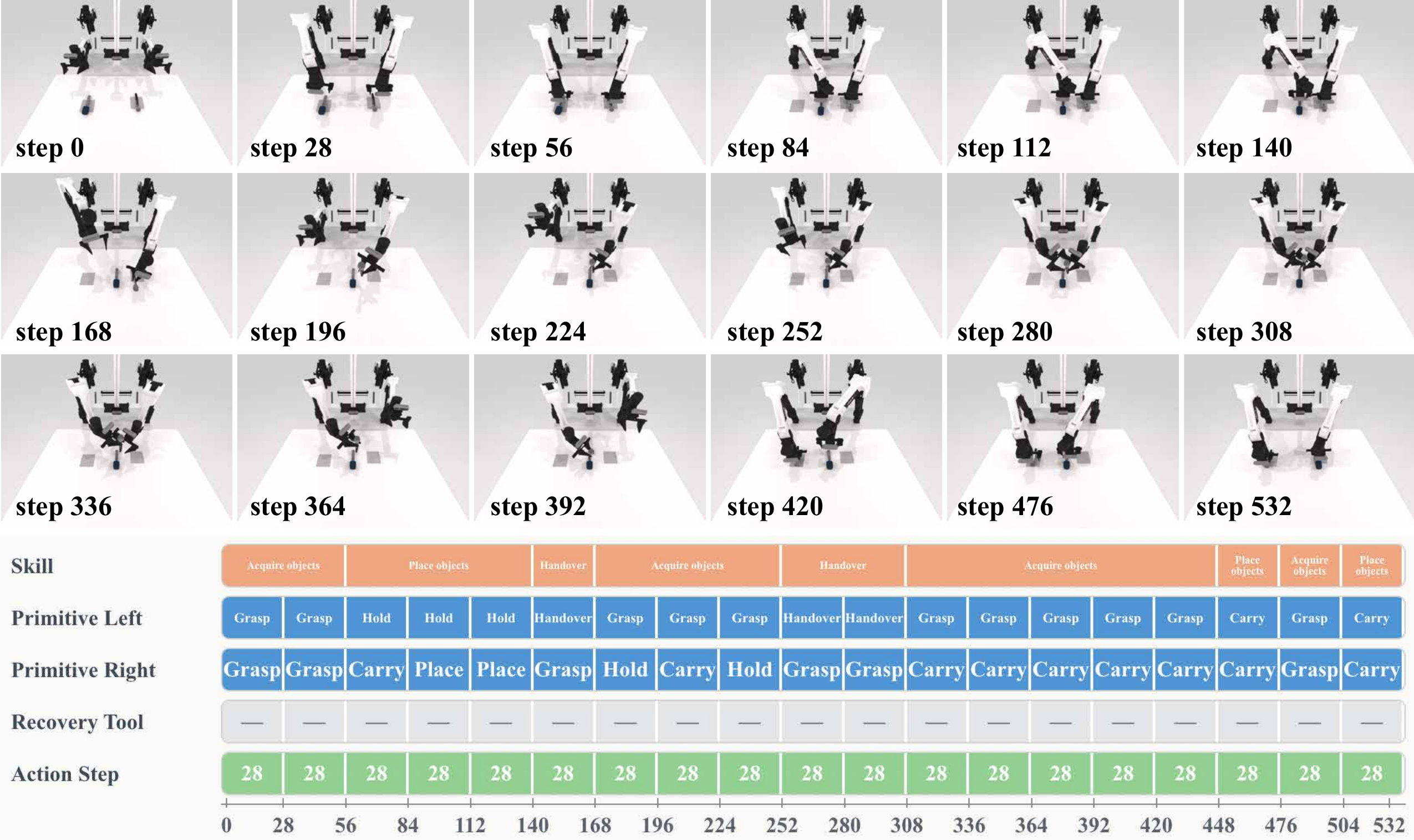}
    \caption{Bimanual microphone exchange. The execution frames and
    timeline show how the arms exchange roles during handovers and
    place the microphones at their target locations.}
    \label{fig:microphone_exchange}
\end{figure*}

\begin{figure*}[t]
    \centering
    \includegraphics[width=0.9\textwidth]{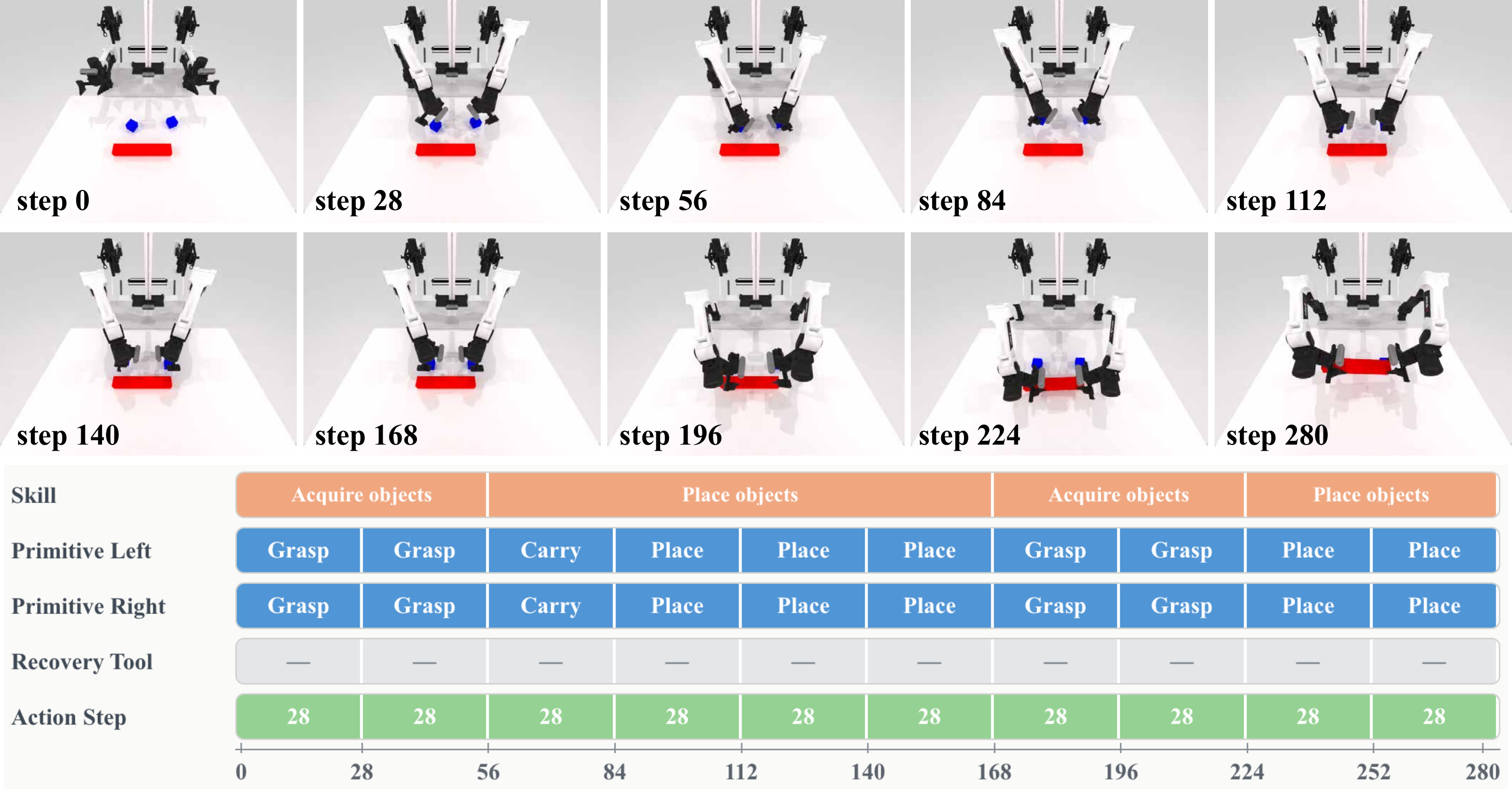}
    \caption{Bimanual bridge building. AWAM positions two support
    blocks before coordinating both arms to place a beam across them.}
    \label{fig:bridge_building}
    \vspace{-12pt}
\end{figure*}

\begin{figure*}[htbp]
    \centering
    \includegraphics[width=0.9\textwidth]{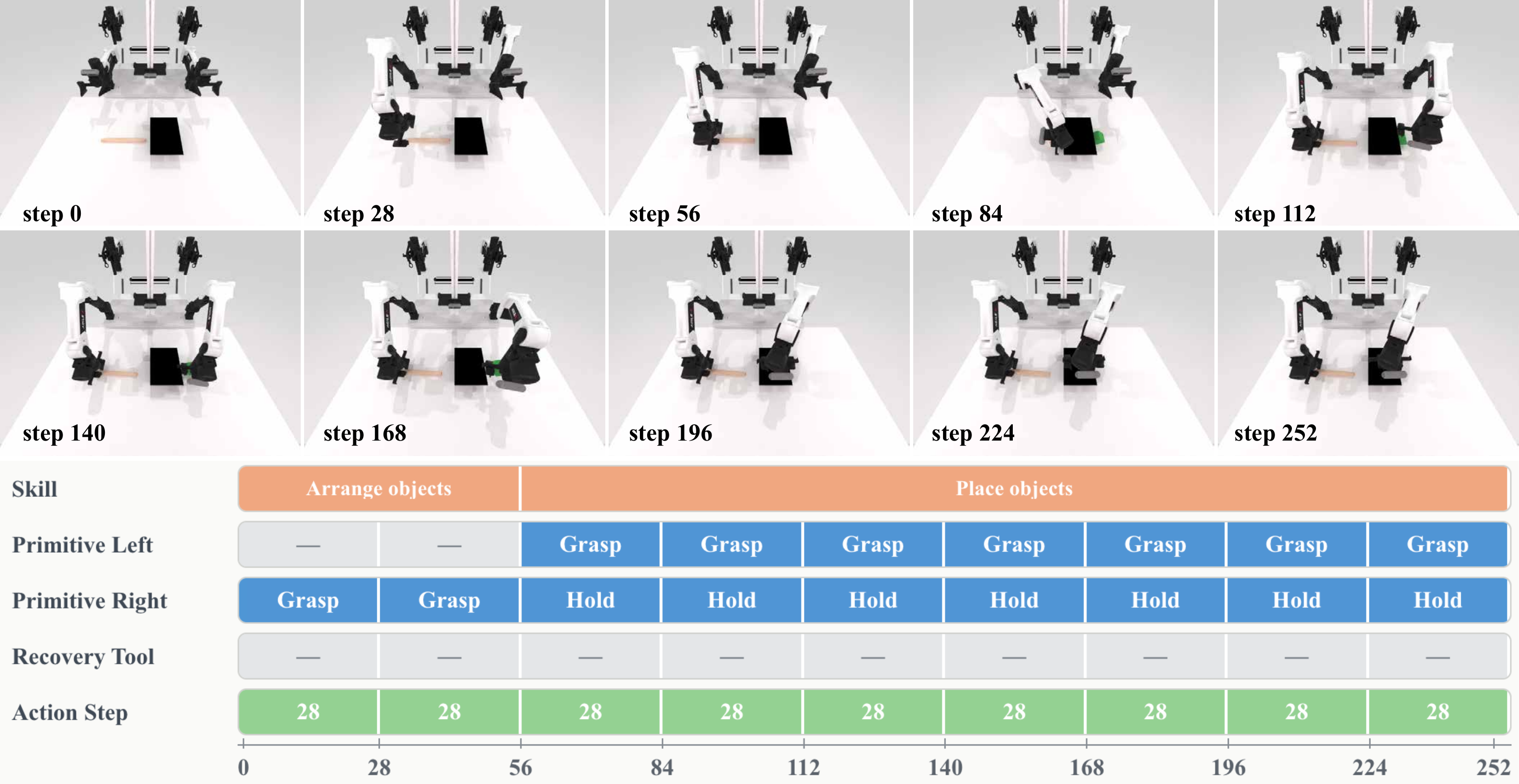}
    \caption{Bimanual block sweeping. One arm holds the dustpan in
    place while the other sweeps the block into it.}
    \label{fig:block_sweeping}
\end{figure*}

\section{Qualitative Results}
\label{app:dual_brain_visualization}

Figures~\ref{fig:bimanual_cooking}--\ref{fig:failure_recovery}
illustrate how AWAM organizes bimanual execution across multi-stage
manipulation tasks. The Agentic Reasoning Brain decomposes task goals
into skills and specifies primitives for each arm, while the
World--Action Brain generates action chunks conditioned on these
primitives and current observations. Each visualization aligns
execution frames with task-level skills, per-arm primitives, and
recovery-tool invocations. This alignment reveals how coordinated
behavior is maintained as the arms change roles, manipulate shared
objects, or respond to an execution failure. The first four examples
show successful runs without recovery-tool invocation, while the
final example illustrates recovery within an ongoing task.

\subsection{Bimanual Cooking}

Figure~\ref{fig:bimanual_cooking} presents a cooking task comprising
object acquisition, food placement, pan shaking, and pouring.
Initially, the arms acquire the food items and the pan. During food
placement, the right arm carries and then holds the pan, while the
left arm manipulates the food items above it. The two arms therefore
perform complementary operations: one maintains the receiving
container, while the other transfers its contents. This coordination
continues across multiple action chunks within the same task-level
skill.

Once the food has been placed in the pan, the right arm transitions
from \texttt{Hold} to \texttt{Shake} and subsequently to
\texttt{Pour}, while the left arm becomes inactive in the primitive
timeline. The execution frames show the pan being tilted over the
plate to transfer its contents. This example illustrates how AWAM
connects successive skills with different coordination requirements,
progressing from simultaneous bimanual manipulation to execution
primarily by one arm while preserving the overall task sequence.

\subsection{Bimanual Microphone Exchange}

Figure~\ref{fig:microphone_exchange} illustrates a microphone
exchange task involving object acquisition, transport, handover,
and placement. Unlike the cooking example, the arms repeatedly
change their operational roles as the microphones move through
the shared workspace. Early in the sequence, the left arm holds
an object while the right arm carries and places another.
The timeline subsequently introduces handover stages, pairing
the left arm's \texttt{Handover} primitive with the right arm's
\texttt{Grasp} primitive.

Following these exchanges, the arms continue grasping, carrying,
and positioning the microphones at their target locations.
The timeline shows that their primitive sequences need not be
identical: one arm can maintain or transfer an object while the
other prepares the next manipulation step. This example highlights
coordination through changing role assignments, where each arm's
operation must remain compatible with the other arm's current
responsibility and the next task-level subgoal.

\subsection{Bimanual Bridge Building}

Figure~\ref{fig:bridge_building} demonstrates sequential
dependencies and synchronized manipulation in bridge construction.
The robot first acquires the two support blocks and places them
at the intended support locations. During this stage, the arms
follow aligned sequences of \texttt{Grasp}, \texttt{Carry}, and
\texttt{Place} primitives, allowing the two supports to be arranged
within the same task phase. Their placement establishes the spatial
configuration required for the subsequent beam-placement stage.

After positioning the supports, both arms return to
\texttt{Grasp} and jointly manipulate the beam before placing it
across the blocks. The execution frames show the transition from
handling separate objects to acting on a shared object, while the
timeline records matching primitive phases for the two arms.
This example illustrates how AWAM combines ordered subgoals with
synchronized bimanual execution: support placement precedes beam
placement, and the final operation requires both arms to contribute
to the same assembly step.

\subsection{Bimanual Block Sweeping}

Figure~\ref{fig:block_sweeping} illustrates sustained complementary
roles during tool-mediated manipulation. The robot first arranges
the tools for sweeping. The right arm then transitions from
\texttt{Grasp} to \texttt{Hold}, maintaining the dustpan in place
while the left arm performs the manipulation needed to sweep the
block into it. The execution frames show one arm maintaining the
receiving tool as the other moves the block toward its opening.

The timeline emphasizes the persistence of this asymmetric
assignment: the right arm remains in \texttt{Hold} across
successive action chunks while the left arm continues manipulating
the sweeping tool. Successful coordination here depends on
maintaining compatible roles throughout the operation, rather
than switching both arms through matching primitive sequences.
This example illustrates how AWAM supports a sustained support
action alongside active manipulation, allowing one arm to provide
the spatial reference needed by the other.

\begin{figure*}[t]
    \centering
    \includegraphics[width=0.9\textwidth]{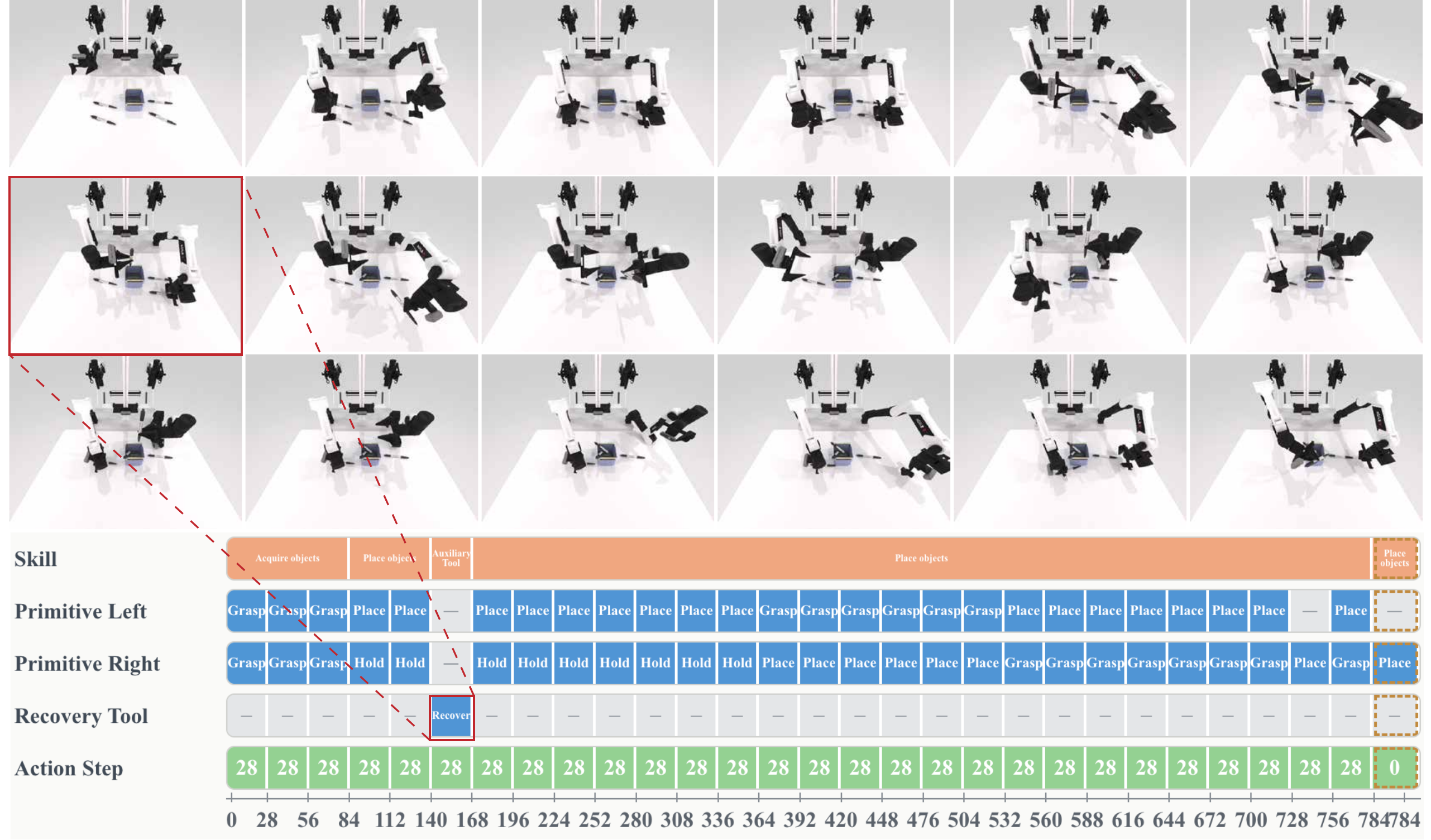}
    \caption{Failure recovery during bimanual pen collection. Red
    annotations link the recovery event in the skill timeline to the
    corresponding execution frame. Subsequent frames show the robot
    resuming pen collection and placement.}
    \label{fig:failure_recovery}
\end{figure*}

\subsection{Recovery from Execution Failures}
\label{app:recovery_visualization}

Figure~\ref{fig:failure_recovery} shows recovery during a bimanual
pen-collection task. The robot initially grasps the pens and begins
placing them into the container, with the left arm executing
\texttt{Place} while the right arm executes \texttt{Hold}.
An execution failure interrupts this progression, after which
the Agentic Reasoning Brain invokes the recovery tool.
The red annotations connect the highlighted execution frame
to the corresponding \texttt{Recover} entry in the timeline.

During the recovery interval, the task-level skill changes to
\texttt{Auxiliary Tool}, and the ordinary per-arm primitive entries
are temporarily inactive. After this intervention, the timeline
returns to object placement, initially restoring the
\texttt{Place}/\texttt{Hold} pairing. Subsequent segments show the
arms alternating between grasping and placing as pen collection
continues.

This example makes the transition from regular manipulation
to recovery and back to task execution explicit. The recovery
operation is inserted into the ongoing sequence while the
overall collection goal is retained. The subsequent execution
frames provide qualitative evidence that AWAM can resume
productive manipulation after a local failure, extending the
nominal coordination behaviors illustrated in the preceding
examples.

\end{document}